\def\year{2021}\relax
\documentclass[letterpaper]{article} 
\usepackage{aaai21}  
\usepackage{times}  
\usepackage{helvet} 
\usepackage{courier}  
\usepackage[hyphens]{url}  
\usepackage{graphicx} 
\usepackage{natbib}  
\usepackage{caption} 
\usepackage{setspace}
\usepackage{amsmath}
\usepackage{amssymb}
\usepackage{graphicx}
\usepackage{comment}
\usepackage{amsmath,amssymb} 
\usepackage{color}
\usepackage{subfigure}
\usepackage{booktabs} 
\usepackage{multirow}
\usepackage{multicol}

\usepackage{algorithm}  
\usepackage{algorithmic}

\graphicspath{{figs/}}

\def\mb{\mathbf}
\def\mbb{\mathbb}
\def\mc{\mathcal}

\def\st{\mbox{s.t. }}
\def\ie{\textit{i.e.}}
\def\wrt{\textit{w.r.t. }}
\def\eg{\textit{e.g.}}
\def\etc{\textit{etc.}}

\title{Multi-Domain Multi-Task Rehearsal for Lifelong Learning}

\author{
    Fan Lyu\textsuperscript{\rm 1}, 
    Shuai Wang\textsuperscript{\rm 1},
    Wei Feng\textsuperscript{\rm 1}\footnote{Corresponding Author.},
    Zihan Ye\textsuperscript{\rm 2},
    Fuyuan Hu\textsuperscript{\rm 2} 
    and 
    Song Wang\textsuperscript{\rm 1, 3}
    \\
}
\affiliations{
    \textsuperscript{\rm 1}Colledge of Intelligence and Computing, Tianjin University\\
    \textsuperscript{\rm 2}School of Electronic \& Information Engineering, Suzhou University of Science and Technology\\
    \textsuperscript{\rm 3}Department of Computer Science and Engineering, University of South Carolina\\
    \{fanlyu, wangshuai201909, wfeng\}@tju.edu.cn, \{zihanye@post, fuyuanhu@mail\}.usts.edu.cn, songwang@cec.sc.edu
}

\begin{document}

\maketitle

\begin{abstract}
Rehearsal, seeking to remind the model by storing old knowledge in lifelong learning, is one of the most effective ways to mitigate catastrophic forgetting, \ie, biased forgetting of previous knowledge when moving to new tasks.
However, the old tasks of the most previous rehearsal-based methods suffer from the unpredictable domain shift when training the new task.
This is because these methods always ignore two significant factors.
First, the Data Imbalance between the new task and old tasks that makes the domain of old tasks prone to shift.
Second, the Task Isolation among all tasks will make the domain shift toward unpredictable directions;
To address the unpredictable domain shift, in this paper, we propose Multi-Domain Multi-Task (MDMT) rehearsal to train the old tasks and new task parallelly and equally to break the isolation among tasks.
Specifically, a two-level angular margin loss is proposed to encourage the intra-class/task compactness and inter-class/task discrepancy, which keeps the model from domain chaos.
In addition, to further address domain shift of the old tasks, we propose an optional episodic distillation loss on the memory to anchor the knowledge for each old task.
Experiments on benchmark datasets validate the proposed approach can effectively mitigate the unpredictable domain shift.
\end{abstract}


\section{Introduction}

Lifelong learning, also known as continual learning and incremental learning, aims to continually learn new knowledge from a sequence of tasks over a lifelong time.
In contrast to traditional supervised learning, the lifelong setting helps machine learning work like a more realistic human learning by acquiring a new skill quickly with new training data.
All the while, \textit{catastrophic forgetting}~\cite{french1999catastrophic,kirkpatrick2017overcoming} is the main challenge for lifelong learning, which happens when the learner forgets the knowledge of old tasks while learning a new task.
To seek a balance between the old tasks and the new task, many methods have been proposed to handle the catastrophic forgetting in recent years.
Following~\cite{de2019continual}, their methods can be categorized into \textit{Rehearsal}~\cite{lopez2017gradient,AGEM,guo2019learning}, \textit{Regularization}~\cite{li2016learning,chaudhry2018riemannian,dhar2019learning} and \textit{Parameter Isolation}~\cite{mallya2018piggyback,yoon2017lifelong}.
Regularization-based and parameter isolation-based methods store no data from old tasks and highly rely on extra regularizers or architectures, resulting in their lower performance than  the rehearsal-based methods.
Rehearsal-based methods store a small number of samples in the training set, the model will retrain the saved data when training the the new task to avoid forgetting.

\begin{figure}[t]
	\centering	
	\subfigure[Traditional rehearsal]{\includegraphics[width=\linewidth]{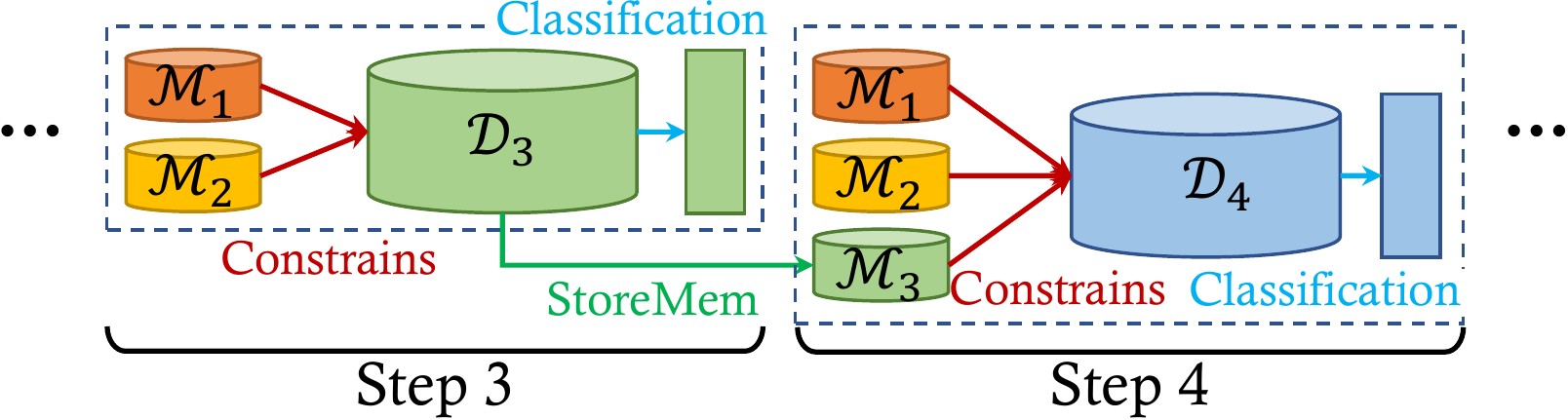}
		\label{fig:rhs}
	}
	\subfigure[MDMT rehearsal]{\includegraphics[width=\linewidth]{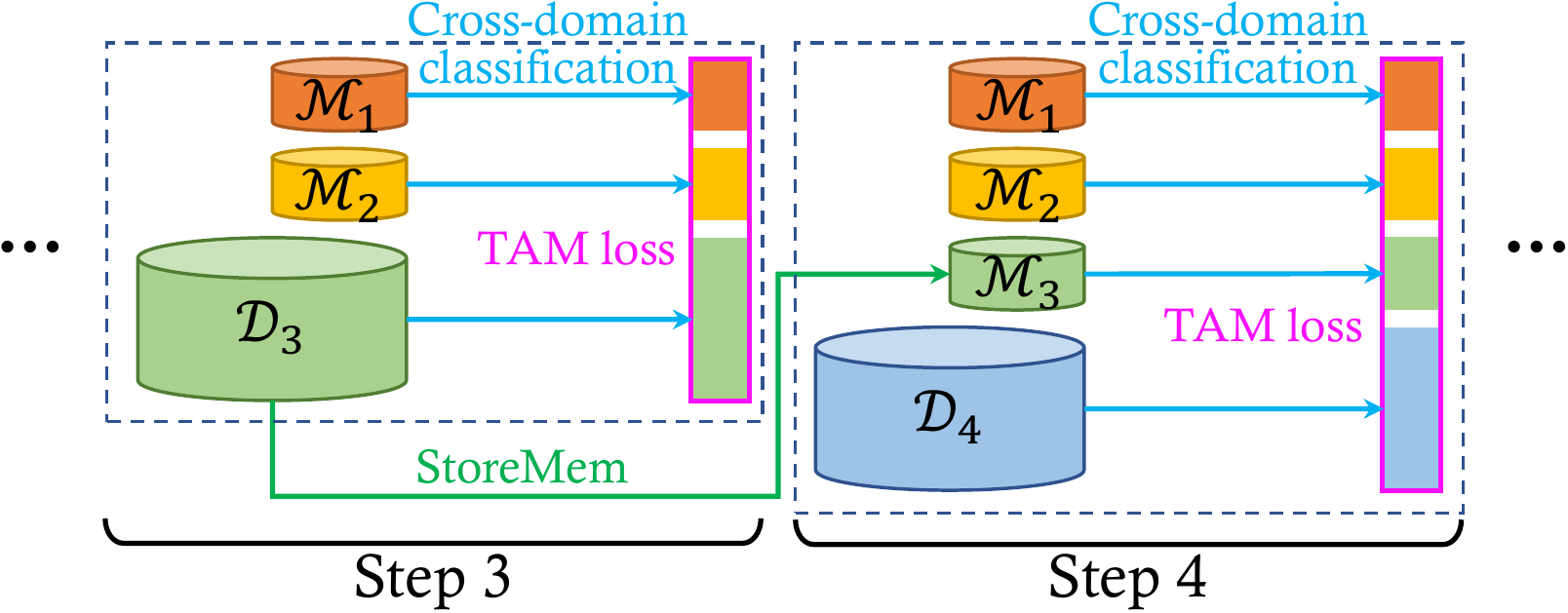}
		\label{fig:mdmtrhs}
	}
	\vspace{-15px}
	\caption{
		(a) Traditional rehearsal-based methods construct single-task learning architecture for the new task (data from training set $\mc{D}$) and treat the old tasks (data from memory $\mc{M}$) as the constraints of its training. 
		(b) The proposed MDMT rehearsal-based method trains old tasks and new task equally and keep tasks from isolation via TAM loss.
	}
	\vspace{-15px}
\end{figure}


At each step of lifelong learning (see Fig.~\ref{fig:rhs}), the most existing rehearsal-based methods~\cite{Rebuffi2016,lopez2017gradient,AGEM,guo2019learning} focus on training the new task while treating the stored data from old tasks as the constraints to preserve their performance.
However, the old tasks in these methods may suffer from \textit{unpredictable domain shift} that arises from two significant factors in the lifelong learning process:
1) The \textit{Data Imbalance} between old and new task.
The shrinkage of training data of old tasks leads to their domains will be prone to shift that manifests as the catastrophic forgetting. 
2) The \textit{Task Isolation} among all tasks (old and new), which makes such domain shift toward unpredictable directions and the boundary between any two tasks may become weak.

%

To address the unpredictable domain shift, in this paper, we propose a Multi-Domain Multi-Task (MDMT) Rehearsal method inspired by the multi-domain multi-task learning~\cite{Yang2014} that considers both multiple tasks \wrt multiple domains and trains them equally.
Specifically, as shown in Fig.~\ref{fig:mdmtrhs}, we first retrain the old tasks along with new task training parallelly rather than setting them as the constraints. 
We separate all these tasks by a Cross-Domain Softmax, which extends the softmax for each isolated task by combining the logits of all other seen tasks and separates them from each other.
Then, to further alleviate the unpredictable domain shift, we propose to leverage a Two-level Angular Margin (TAM) loss to encourage the intra-class/task compactness and the inter-class/task discrepancy on the basis of Cross-Domain Softmax.
In addition, we present an optional Episodic Distillation (ED) loss on all buffer memories for old tasks that suppress the domain shift by storing the latent representations of each sample in memories.
We evaluate our MDMT rehearsal on four popular lifelong learning datasets for image classification and achieve new state-of-the-art performance. The experimental results show the proposed MDMT rehearsal can significantly mitigate the unpredictable domain shift.
Our contributions are three-fold:
(1) We propose a Multi-Domain Multi-Task Rehearsal method for lifelong learning, which parallelly and equally trains the old and new tasks and separate them by a Cross-Domain Softmax function.
(2) We propose a Two-level Angular Margin (TAM) loss for lifelong learning to further boost the Cross-Domain Softmax for the sake of intra-class/task compactness and the inter-class/task discrepancy.
(3) We build an optional Episodic Distillation loss to reduce the domain shift in lifelong progress.

\section{Related Work}

\noindent
\textbf{Lifelong learning}.
In contrast to static machine learning~\cite{he2016deep,deng2018visual,lyu2019attend,lyu2020vtgraphnet}, Lifelong Learning~\cite{ring1997child,thrun1998lifelong} is proposed to improve the self-learning ability of the machine that continually learns new knowledge.
The previous solutions to the catastrophic forgetting~\cite{french1999catastrophic,kirkpatrick2017overcoming} have been proposed in recent years and can be categorized into \textit{regularization-based}, \textit{parameter isolation-based} and \textit{rehearsal-based} methods~\cite{de2019continual}.
Regularization-based methods~\cite{li2016learning,chaudhry2018riemannian,dhar2019learning} store no data but propose to use extra regularization terms in the loss function to consolidate previous knowledge.
Parameter isolation-based methods~\cite{mallya2018piggyback,yoon2017lifelong} freeze the task-specific parameters and grow new branches for new tasks to bring in new knowledge.
Although many progress have been made on regularization-based and parameter isolation-based methods, their performance still has a big gap with rehearsal-based methods. 
Rehearsal-based methods store some knowledge of old tasks to remind the model.
According to different saved forms, existing methods can be categorized into three groups:
(1) by saving the raw data (Rehearsal, \eg, image)~\cite{Rebuffi2016,lopez2017gradient,AGEM,guo2019learning}, the model can retrain the saved data with the training on the new task together, and these methods often construct single objective for the new task and set the saved data from old tasks as the constrains;
(2) by saving the latent features for selected samples (Latent-rehearsal)~\cite{Pellegrini2019}, the model slows down learning at the layers below the rehearsal layer and leaves the layers above free to learn at full pace;
(3) by building generative model to synthesize data (Pseudo-rehearsal)~\cite{Shen2020,Ven2018,Lesort2018}, the knowledge can be saved as parameters rather than data.
In this paper, we only consider the native rehearsal by storing raw data in image classification.

\noindent
\textbf{Multi-domain multi-task learning.}
Multi-domain learning~\cite{Nam2015,Tang2020} and multi-task learning~\cite{lin2019pareto,sener2018multi} have been studied for years. 
Multi-domain learning refers to sharing information about the same problem across different contextual domains, while multi-task learning addresses sharing information about different problems in the same domain. 
By considering both multiple domains and multiple tasks, Multi-domain multi-task (MDMT) learning was first proposed in \cite{Yang2014}, and has been applied to classification~\cite{Peng2016} and semantic segmentation~\cite{Fourure2017}, \etc.
The common method for MDMT learning is to construct parallel data streams and seek to build the correlations among tasks.
Here, we explain why we decide to formulate the lifelong learning problem into a MDMT learning problem.
(1) By storing some samples of a task into a memory, MDMT learning can significantly train them together, which helps mitigate the task isolation in the traditional rehearsal-based lifelong learning. 
(2) MDMT learning can help suspending the domain shift to some extent by making classifiers perceive each other.

\noindent
\textbf{Margin loss and distillation loss.}
The margin based Softmax explicitly adds a margin to each logit to improve feature discrimination. 
L-Softmax~\cite{liu2016large-margin} and SphereFace~\cite{DBLP:conf/cvpr/LiuWYLRS17} add multiplicative angular margin to squeeze each class. CosFace~\cite{DBLP:conf/cvpr/WangWZJGZL018,DBLP:journals/spl/WangCLL18} and ArcFace~\cite{deng2018arcface} add additive cosine margin and angular margin, respectively, for easier optimization. 
Based on ArcFace, we propose a Two-level Angular Margin loss to guarantee both inter-class/task compactness and intra-class/task discrepancy.
The knowledge distillation~\cite{Hinton2015} from the teacher network to the student network transfers the knowledge about smoothed probability distribution of the output layer of the teacher network to the student network. 
Inspired by this, we propose to build distillation loss between the old and new models on old tasks by storing the latent representation of stored data.

\begin{figure*}[t]
	\centering
	\includegraphics[width=.85\linewidth]{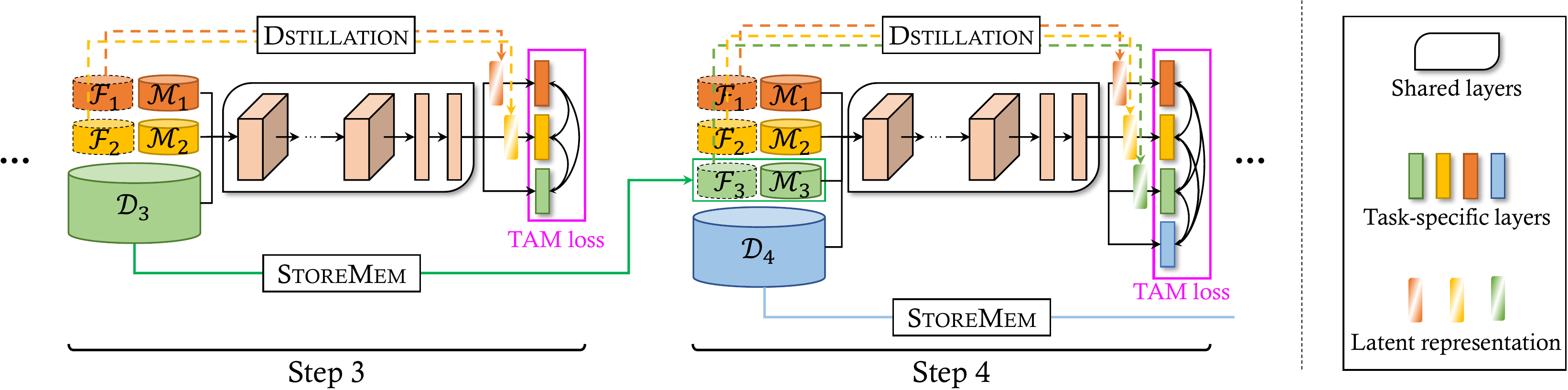}
	\caption{
		Training procedure of the proposed MDMT rehearsal based lifelong learning.
		At each step, a small number of samples will be saved into memory $\mc{M}$ and the corresponding latent representations will be saved into $\mc{F}$.
		TAM loss guarantee the intra-class/task compactness and inter-class/task discrepancy.
		Episodic Distillation loss helps further to reduce the domain shift of the old tasks.
		The dashed elements mean the optional operation.
	}
	\label{fig:method}
	\vspace{-20px}
\end{figure*}

\section{Methodology}

\renewcommand{\algorithmicrequire}{ \textbf{Procedure}}     

\subsection{Multi-domain multi-task rehearsal}

Suppose there are $T$ different tasks with respect to datasets $\{\mc{D}_1,\cdots,\mc{D}_T\}$. 
For the $t$-th dataset (task), $\mc{D}_t=\{(x_{t,1},y_{t,1}),\cdots,(x_{t,N_t},y_{t,N_t})\}$, where $x_{t,i}\in\mc{X}_t$ is the $i$-th input data, $y_{t,i}\in\mc{Y}_t$ is the corresponding label and $N_t$ is the number of samples.
$\mc{D}_t$ can be split into a training set $\mc{D}_t^{\text{trn}}$ and a testing set $\mc{D}_t^{\text{tst}}$, and we denote $\mc{D}_t$ as $\mc{D}_t^{\text{trn}}$ in our presentation for simple denotation.
Lifelong learning aims at learning a predictor $f_t:\mc{X}_k\rightarrow\mc{Y}_k,~k\in\{1,\cdots,t\}$, which can predict tasks that have been learned at any time.
The rehearsal-based lifelong learning~\cite{Rebuffi2016,lopez2017gradient,riemer2018learning,AGEM,guo2019learning} builds a memory buffer $\mc{M}_k\subset\mc{D}_k$ with small-size for each previous task $k$, \ie, $|\mc{M}_k|\ll|\mc{D}_k|$. 
Following~\cite{lopez2017gradient}, when training a task $t\in\{1,\cdots,T\}$, for all $\mc{M}_k$ that $k<t$, the rehearsal-based lifelong learning can be modeled as a single objective optimizing problem: 
\begin{equation}
	\begin{aligned}
		\mathop{\arg\min}_{\theta,\theta_t}\quad&\ell(f_\theta, f_{\theta_t}, \mc{D}_t),\quad\\
		\st\quad&\ell(f_\theta, f_{\theta_k},\mc{M}_k)\le\ell(f_\theta^{t-1},f^{t-1}_{\theta_k}, \mc{M}_k),\quad\forall k<t,
	\end{aligned}
	\label{eq:eps}
\end{equation}
where $\ell$ is the empirical loss. $\theta$ is the shared parameter across all tasks while $\theta_k$ and $\theta_t$ are the task-specific parameters. 
The constraints above are designed to prevent the performance degradation of previous tasks.
Then, the problem can be reduced to find an optimal gradient that benefits all tasks.
To inspect the increase in old tasks' loss, \cite{lopez2017gradient,AGEM,guo2019learning} compute the angle between the gradient of each old task and the proposed gradient update on the current task. 


However, such a single objective optimization on the current task for rehearsal-based lifelong learning over-emphasizes the new task while ignoring the difference among tasks.
In other words, the old tasks can only play the role of source domain to be transferred into the current training model.
The domain of old tasks will significantly shift because of the rectified gradient that the gradient norm of new task is much larger than the old tasks', which may induce the domain overlap.

\begin{algorithm}[tb]
	\caption{MDMT rehearsal based lifelong learning.}
	\label{alg:method}
	\begin{algorithmic}
		\REQUIRE \textsc{Train}($f_\theta$, $f_{\theta_{1:T}}$, $\{\mc{D}_1^{\text{trn}},\cdots,\mc{D}_T^{\text{trn}}\}$)
		
		\STATE $\mc{M},\mc{F}\leftarrow\{\}$
		\FOR{$t=1$ {\bfseries to} $T$}
		\FOR{$(\mb{x},y)\in\mc{D}_t^{\text{trn}}$}
		\STATE $g,g_1\leftarrow\nabla_\theta\ell(f_\theta(\mb{x},t),y)$
		\IF{$t=1$}
		\STATE $\tilde{g}\leftarrow g$
		\ELSE
		\STATE $g^\text{ref},g_{1:t-1}\leftarrow\nabla_\theta\ell(f_\theta, f_{\theta_{1:t-1}}, \mc{M})$
		\STATE $g^\text{ref}\leftarrow g^\text{ref} + \nabla_\theta\tilde{\ell}(f_\theta, \mc{F}_\text{ref})$
		\STATE $\tilde{g}\leftarrow g+g^{\text{ref}}$
		\ENDIF		
		\STATE $\theta\leftarrow\theta-\text{StepSize}\cdot\tilde{g}$
		\STATE $\theta_{1:t}\leftarrow\theta_{1:t}-\text{StepSize}\cdot g_{1:t}$
		\ENDFOR
		\STATE $\mc{M},\mc{F}\leftarrow\text{\textsc{StoreMem}}(\mc{M},\mc{F},\mc{D}_t^{\text{trn}},f_\theta)$
		\ENDFOR

		~

		%

		\REQUIRE {\textsc{StoreMem}}($\mc{M}$, $\mc{F}$, $\mc{D}$, $f$)
		\FOR{$i=1$ {\bfseries to} $|\mc{M}|/T$}
		\STATE $(\mb{x},y)\sim\mc{D}$
		\STATE $\mc{M}\leftarrow\mc{M}+(\mb{x},y)$
		\STATE $\mc{F}\leftarrow\mc{F}+ f(\mb{x})$
		\ENDFOR
		\STATE {\bfseries Return} $\mc{M}$, $\mc{F}$

		~
		
		\REQUIRE \textsc{Eval}($f_\theta$, $f_{\theta_{1:T}}$, $\{\mc{D}_1^{\text{tst}},\cdots,\mc{D}_T^{\text{tst}}\}$)
		\STATE $a\leftarrow0\in\mbb{R}^T$
		\FOR{$t=1$ {\bfseries to} $T$}
		\STATE $a_t\leftarrow0$
		\FOR{$(\mb{x},y)\in\mc{D}_t^{\text{tst}}$}
		\STATE $a_t\leftarrow a_t+\text{Accuracy}(f_{\theta_t}(f_\theta(\mb{x},t)),y)$
		\ENDFOR
		\STATE $a_t\leftarrow a_t/|\mc{D}_t^{\text{tst}}|$
		\ENDFOR
		\STATE {\bfseries Return} $a$
	\end{algorithmic}
\end{algorithm}

In contrast, this paper treats the problem as a Multi-Domain Multi-Task (MDMT) learning problem to jointly and equally improve the current task as well as the old tasks:
\begin{equation}
	\begin{aligned}
		\mathop{\arg}_{\theta,\{\theta_1,\cdots,\theta_t\}}\quad&\{\min\ell(f_\theta,  f_{\theta_t}, \mc{D}_t), \min\ell(f_\theta,  f_{\theta_{k}},\mc{M}_{k}),\cdots,\\
		&~~~~~\min\ell(f_\theta, f_{\theta_{1}}, \mc{M}_1)\},\\
		\st\quad&d(f_i,f_j) \ge d(f^{t-1}_i,f^{t-1}_j), i,j\in[1,t], i\ne j,
	\end{aligned}
	\label{eq:mdmt}
\end{equation}
where $f_i=f_\theta(\mc{D}_i)$ if $i=t$ and $f_i=f_\theta(\mc{M}_i)$ if $i<t$.
$d$ means the distance between two domains.
For the $t$ tasks w.r.t. datasets $\{\mc{D}_1,\cdots,\mc{D}_t\}$, a MDMT rehearsal model trains $t$ tasks parallelly and equally.
The constraints above mean the domain distance between any two tasks should not be smaller than the model trained on the last task.
Note that we only consider the situation that the tasks are irrelevant as the common lifelong learning.

We make two key operations to solve the Eq.\eqref{eq:mdmt} efficiently.
First, we transform the multi-objective optimization as a single-objective optimization problem by ensembling all these objectives as the traditional solution to multi-task learning~\cite{lin2019pareto,sener2018multi}.
\begin{equation}
	\mathop{\arg\min}_{\theta}\quad\ell(f_\theta,f_{\theta_{t}}, \mc{D}_t)+\sum_{k=1}^{t-1}\ell(f_\theta,f_{\theta_{k}}, \mc{M}_k),
	\label{eq:mdmt_2}
\end{equation}
Second, it exists high memory-cost to calculate the distance between any two domains and store old predictors $f_\theta^{t-1}$, but we can do this in a simple yet effective way by extending the softmax function for each task as
\begin{equation}
	\ell_k=-\frac{1}{N_k} \sum_{n=1}^{N_k} \log \frac{e^{\left(W^k_{y_{n}}\right)^{T} x_{n}+b_{y_{n}}}}{\sigma_n},
	\label{eq:softmax}
\end{equation}
where 
\begin{equation}
	\sigma_n=\sum_{j=1}^{C_k} e^{\left(W^k_{j}\right)^{T} x_{n}+b_{j}}+\sum_{i=1,i\ne k}^{t}\sum_{j=1}^{C_i} e^{\left(W^i_{j}\right)^{T} x_{n}+b_{j}}.
	\label{eq:softmax-g}
\end{equation}
$N_k$ is the batch size for task $k$ and $W^k_{j} \in \mathbb{R}^{d}$ denotes the $j$-th column of the weight $W^k \in \mathbb{R}^{d\times C_k}$ in the last fully-connected layer for task $k$ and $C_k$ is the class number.
We name this extension as Cross-Domain Softmax (CDS), which combines the logits from other classifiers and is similar to a native softmax to a classification problem with total $\sum_{k=1}^{t}C_k$ class.
Here, we discuss the difference. 
For MDMT rehearsal, different tasks never share a same classifiers as common classification, \ie, the classifiers for different tasks lack mutual perception. 
By combining the logits form other tasks, the tasks can perceive and separate from each other.
The previous methods update the model by the optimal gradient that highly rely on the angle between the gradients of old and new tasks. 
In contrast, we directly obtain the hybrid gradient for the shared layers by ensembling the gradients from the new task and old tasks as
$\tilde{g}\leftarrow\sum_{k=1}^t g_k$.


We compare our MDMT rehearsal with some famous rehearsal-based lifelong works:

\begin{figure}[t]
	\centering	
	\includegraphics[width=\linewidth]{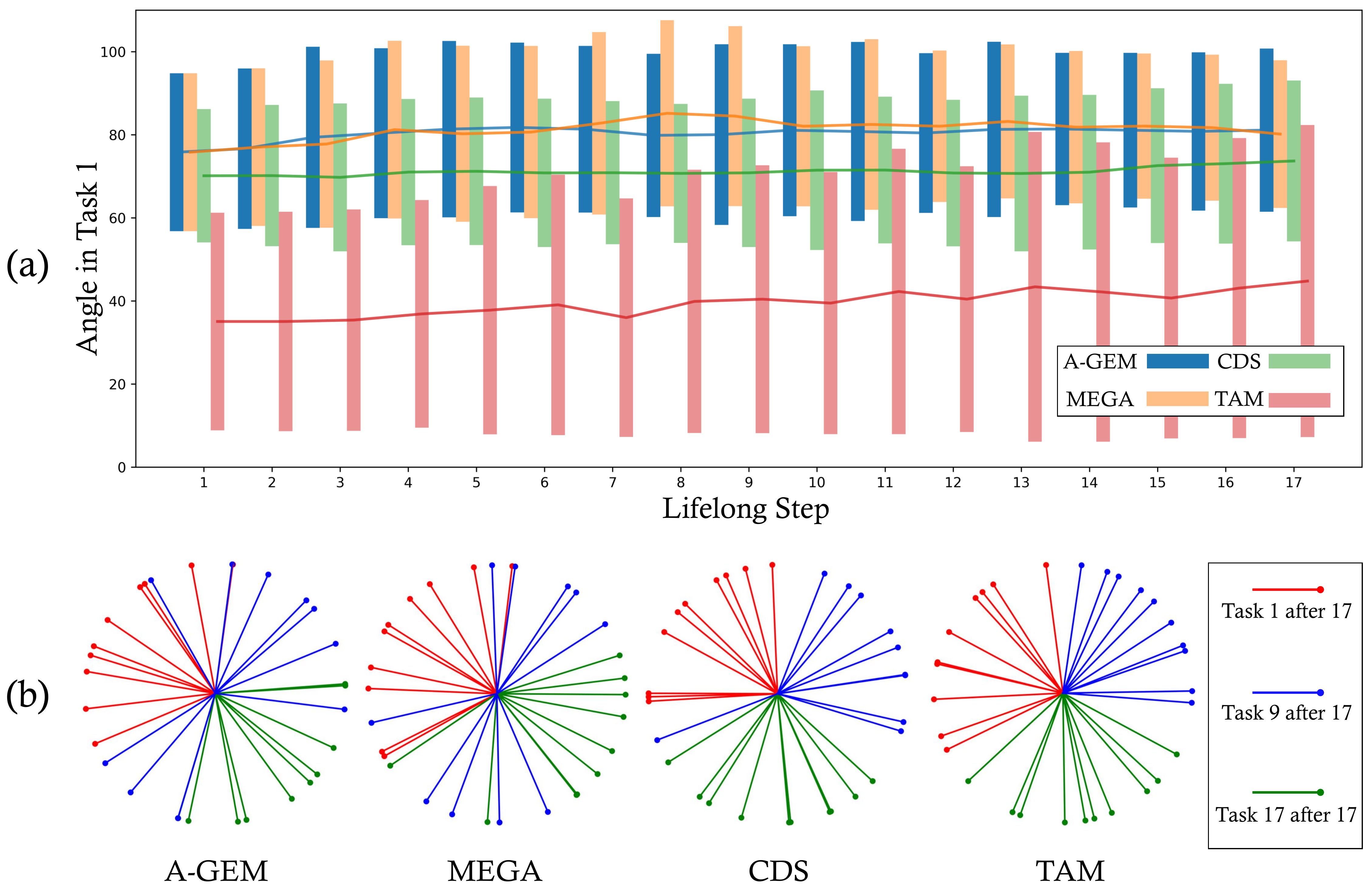}
	\caption{On Permuted MNIST, (a) the changes of angle range between feature and the target weight center of task 1 along the lifelong learning; (b) the angulars relations of class centers of task 1, 9 and 17 after trained on task 17.}
	\label{fig:angular}
	\vspace{-15px}
\end{figure}

\noindent
\textbf{iCaRL}~\cite{Rebuffi2016} saves small number of samples to make the model not to forget old class, but they classify samples by the nearest prototype, which is not suitable for task-incremental lifelong learning because the task-specific parameters are ignored.

\noindent
\textbf{GEM/A-GEM}~\cite{lopez2017gradient,AGEM} propose to solve forgetting by finding the optimal gradient that saves the old tasks from being corrupt, and they focus on training the new task with single objective optimization while ignore the domain shift of old tasks.

\noindent
\textbf{ER}~\cite{chaudhry2019on} extends Experience Replay~\cite{rolnick2018experience} for reinforcement lifelong learning and be proven better than A-GEM. However, they never consider the relations among all tasks, which makes the domains of old task may significantly shift.

\noindent
\textbf{PRD}~\cite{hou2018lifelong} proposes to treat lifelong learning as a multi-task learning problem and proposes to build a distillation module with one saved CNN expert as teacher for each old task.
Differently, we would like to build a MDMT rehearsal that leverage the expanded softmax without saving many extra models.

\subsection{Two-level angular margin loss}

The proposed MDMT rehearsal helps to jointly and equally train the new task and retrain the old tasks, making all tasks perceive each other. 
Nonetheless, the softmax loss is not efficient enough because it does not explicitly encourage intra-class compactness and inter-class discrepancy, in coping with which, large margin based softmax is widely used in recent discriminative problems~\cite{deng2018arcface,liu2016large-margin}.
However, these methods cannot be directly applied to MDMT rehearsal based lifelong learning because these methods place the large margin only to single task and can not be applied to multiple tasks scenario.


In this paper, we propose two levels margin, \ie, \textit{class level} and \textit{task level}, on softmax for each task (Eq.~\eqref{eq:softmax}).
Our work is based on the popular large margin based softmax method Arcface~\cite{deng2018arcface} where the large margin is added to the angle between weight and feature, which has been proven effective and efficient.
Specifically, Arcface deletes the bias and transforms the logit fed into the softmax as $W_{j}^{T} x_{i}=\left\|W_{j}\right\|\left\|x_{i}\right\| \cos \theta_{j}$, where $\theta_{j}$ is the angle between the weight $W_{j}$ and the feature $x_i$, then an angular margin $m$ is placed between different classes 
\begin{equation}
	\ell=-\frac{1}{N} \sum_{i=1}^{N} \log \frac{e^{s\cdot\cos \left(\theta_{y_{i}}+m\right)}}{e^{s\cdot\cos \left(\theta_{y_{i}}+m\right)}+\sum_{j=1, j \neq y_{i}}^{n} e^{s\cdot \cos \theta_{j}}},
	\label{eq:arcface}
\end{equation}
where the individual weight $||W_j||$ is fixed to $1$ by $l_2$ normalization and the embedding feature $||x_i||$ is fixed to $s$ by $l_2$ normalization and rescale.
The normalization on features and weights makes the predictions only depend on the angle between them. 
Such a geodesic distance margin between the sample and centers makes the prediction gain more intra-class compactness and inter-class discrepancy.

Based on Eq.~\eqref{eq:arcface}, we propose our Two-level Angular Margin (TAM) loss for the task $k\in[1,t]$ 
\begin{equation}
	\ell_{k}=-\frac{1}{N_k} \sum_{n=1}^{N_k} \log \frac{e^{s\cdot\cos \left((\theta^k_{y_{n}}+m^\text{c})+m^\text{t}\right)}}{\sigma_n},
	\label{eq:dmsl}
\end{equation}
where 
\begin{equation}
	\begin{aligned}
		\sigma_n = &e^{s\cdot\cos \left((\theta_{y_{i}}^k+m^\text{c})+m^\text{t}\right)} + \sum_{j=1, j \neq y_{i}}^{C_k} e^{s\cdot \cos (\theta_{j}^k+m^\text{t})}+\\
		&\sum_{i=1, i \neq k}^{t}\sum_{j=1}^{C_i} e^{s\cdot \cos \theta_{j}^i}.
	\end{aligned}
\end{equation}
In Eq.~\eqref{eq:dmsl}, we add class-level margin $m^\text{c}$ and task-level margin $m^\text{t}$ on the angular.
$m^\text{c}$ is similar to $m$ in Eq.~\eqref{eq:arcface}, which controls the intra-task class compactness and discrepancy and be proven effective on discriminative learning~\cite{deng2018arcface}.
$m^\text{t}$ controls the task compactness and discrepancy, which ensures the knowledge of each task not to mix up with others.

As shown in Fig.~\ref{fig:angular}, the proposed TAM loss produces two advantages for MDMT rehearsal based lifelong learning.
First, TAM helps the model to better discriminate into a task.
Although the CDS has a better angle between feature and its target weight, TAM loss even reduce the angle to a smaller than CDS, which expresses the effect of $m_\text{c}$.
Second, TAM loss mitigates the domain overlap caused by the domain shift by forcing tasks to separate.
We can also see that for the angles among weights center, TAM loss can significantly separate old and new tasks, which expresses the effect of $m_\text{t}$.
However, it is still difficult to omit the domain shift because of the extreme data imbalance between old tasks and new task.
Thus, we construct an optional Episodic distillation loss for the MDMT rehearsal based lifelong process.

\begin{table*}[t]
	\centering
	\caption{Comparison with different state-of-the-arts. The numbers are averaged across 5 runs using a different seed each time.}
	\vspace{-10px}
	\label{tab:sota}
	\resizebox{.99\linewidth}{!}{
		\begin{tabular}{l|cccc|cccc}
			\toprule
			\multirow{2}{*}{Method}  & \multicolumn{4}{c|}{\textbf{Permuted MNIST}} & \multicolumn{4}{c}{\textbf{Split CIFAR}}  \\
			& {$A_{\text{T}}(\%)$} & {$F_{\text{T}}$} & {$LCA_{10}$} & {$LTR$} & {$A_{\text{T}}(\%)$} & {$F_{\text{T}}$} & {$LCA_{10}$} & {$LTR$} \\
			\hline
			Joint   	& $95.30$ & - & - & - & $68.30$ & - & - & -  \\\hline
			VAN    	& $47.55\pm2.37$ & $0.52\pm0.026$ & $0.259\pm0.005$ & $5.375\pm0.194$ & $40.44\pm1.02$ & $0.27\pm0.006$ & $0.309\pm0.011$ & $2.613\pm0.174$ \\
			EWC 	& $68.68\pm0.98$ & $0.28\pm0.010$ & $0.276\pm0.002$ & $3.292\pm0.135$ & $42.67\pm4.24$ & $0.26\pm0.039$ & $0.336\pm0.010$ & $2.493\pm0.427$ \\
			MAS    	& $70.30\pm1.67$ & $0.26\pm0.018$ & $0.298\pm0.006$ & - & $42.35\pm3.52$ & $0.26\pm0.030$ & $0.332\pm0.010$ & - \\
			RWalk   & $85.60\pm0.71$ & $0.08\pm0.007$ & $\mathbf{0.319}\pm0.003$ & - & $42.11\pm3.69$ & $0.27\pm0.032$ & $0.334\pm0.012$ & - \\
			MER     & - 			 & - 			  & -               & - & $37.27\pm1.68$ & $\mathbf{0.03}\pm0.030$ & $0.051\pm0.101$ & - \\
			GEM     & $89.50\pm0.48$ & $0.06\pm0.004$ & $0.230\pm0.005$ & - & $61.20\pm0.78$ & $0.06\pm0.007$ & $0.360\pm0.007$ & - \\
			A-GEM   & $89.32\pm0.46$ & $0.07\pm0.004$ & $0.277\pm0.008$ & $0.716\pm0.048$ & $61.28\pm1.88$ & $0.09\pm0.018$ & $0.350 \pm0.013$ & $0.643\pm0.124$ \\
			ER   	& $90.47\pm0.14$ & $0.03\pm0.001$ & $0.184\pm0.004$ & $0.367\pm0.013$ & $63.97\pm1.30$ & $0.06\pm0.006$ & $0.349 \pm0.105$ & $0.451\pm0.333$ \\
			MEGA   	& $91.21\pm0.10$ & $0.05\pm0.001$ & $0.283\pm0.004$ & $0.524\pm0.017$ & $66.12\pm1.94$ & $0.06\pm0.015$ & $\mathbf{0.375}\pm0.012$ & $0.356\pm0.114$ \\\hline
			MDMT-R   	& $\mb{94.33}\pm0.04$ & $\mb{0.02}\pm0.000$ & $0.298\pm0.003$ & $\mb{0.247}\pm0.009$ & $\textbf{69.20}\pm1.60$ & $0.04\pm0.010$ & $0.334\pm0.008$ & $\mb{0.283}\pm0.099$ \\\midrule
			\multirow{2}{*}{Method}  & \multicolumn{4}{c|}{\textbf{Split CUB}} & \multicolumn{4}{c}{\textbf{Split AWA}}  \\
			& {$A_{\text{T}}(\%)$} & {$F_{\text{T}}$} & {$LCA_{10}$} & {$LTR$} & {$A_{\text{T}}(\%)$} & {$F_{\text{T}}$} & {$LCA_{10}$} & {$LTR$} \\
			\hline
			Joint   	& $65.60$ & - & - & - & $64.80$ & - & - & -  \\\hline
			VAN    	& $53.89\pm2.00$ & $0.13\pm0.020$ & $0.292\pm0.008$ & $0.976\pm0.215$ & $30.35\pm2.81$ & $0.04\pm0.013$ & $0.214\pm0.008$ & $0.202\pm0.090$ \\
			EWC 	& $53.56\pm1.67$ & $0.14\pm0.024$ & $0.292\pm0.009$ & $1.021\pm0.210$ & $33.43\pm3.07$ & $0.08\pm0.021$ & $0.257\pm0.011$ & $0.675\pm0.214$ \\
			MAS    	& $54.12\pm1.72$ & $0.13\pm0.013$ & $0.293\pm0.008$ & - & $33.83\pm2.99$ & $0.08\pm0.022$ & $0.257\pm0.011$ & - \\
			RWalk   & $54.11\pm1.71$ & $0.13\pm0.013$ & $0.293\pm0.009$ & - & $33.63\pm2.64$ & $0.08\pm0.023$ & $0.258\pm0.011$ & - \\
			PI 		& $55.04\pm3.05$ & $0.12\pm0.026$ & $0.292\pm0.010$ & - & $33.86\pm2.77$ & $0.08\pm0.022$ & $0.259\pm0.011$ & - \\
			A-GEM   & $61.82\pm3.72$ & $0.08\pm0.021$ & $0.302\pm0.011$ & $0.456\pm0.174$ & $44.95\pm2.97$ & $0.05\pm0.014$ & $0.287\pm0.012$ & $0.178\pm0.082$ \\
			ER   	& $73.63\pm0.52$ & $\mb{0.01}\pm0.005$ & $0.265\pm0.004$ & $\mb{0.001}\pm0.001$ & $54.27\pm4.05$ & $\mb{0.02}\pm0.030$ & $0.293\pm0.009$ & $0.014\pm0.015$ \\
			MEGA   	& $80.58\pm1.94$ & $\mb{0.01}\pm0.017$ & $0.311\pm0.010$ & ${0.002}\pm0.002$ & $54.28\pm4.84$ & $0.05\pm0.040$ & $\mb{0.305}\pm0.015$ & $0.070\pm0.114$ \\\hline
			MDMT-R   	& $\textbf{84.27}\pm1.63$ & $\mb{0.01}\pm0.015$ & $\mb{0.337}\pm0.013$ & $0.017\pm0.014$ & $\textbf{61.56}\pm3.36$ & $\mb{0.02}\pm0.027$ & $0.298\pm0.008$ & $\mb{0.002}\pm0.002$ \\
			\bottomrule
	\end{tabular}}
	\vspace{-10px}
\end{table*}

\subsection{Episodic distillation}

In this paper, we propose a simple yet effective solution to further mitigate the domain shift for old task named Episodic Distilllation (ED) loss.
The main role the ED loss played is to reduce the feature distribution change along with the lifelong process as far as possible.
First, apart from the sampled training data stored in memory, \ie, $\mc{M}_k=\{(x_{k,1},y_{k,1}),\cdots,(x_{k,|M_k|},y_{k,|M_k|})\}\subset\mc{D}_k$,
we also store the corresponding latent representations when they are first trained, denoted as $\mc{F}_k=\{\mb{f}_{k,1},\cdots,\mb{f}_{k,|M_k|}\}$.
Then, we train the model with an updated objective:
\begin{equation}
	\mathop{\arg\min}_{\theta}\quad\ell(f_\theta, f_{\theta_t}, \mc{D}_t)+
	\sum_{k=1}^{t-1}\left[\ell(f_\theta, f_{\theta_k}, \mc{M}_k)+\tilde{\ell}(f_\theta, \mc{F}_k)\right],
	\label{eq:finalloss}
\end{equation}
where
\begin{equation}
	\tilde{\ell}(f_\theta, \mc{F}_k)\triangleq\frac{1}{N_k}\sum_i\tilde{\ell}_i(f_\theta(\mb{x}_{k,i}), \mb{f}_{k,i}).
	\label{eq:lossf}
\end{equation}
$\tilde{\ell}_i$ is the ED loss that can be in many formats, and we choose the Mean Square Error (MSE). 
By training with Eq.~\eqref{eq:finalloss} in each step, we can ease the shift effectively.

ED loss is an optional loss function and builds extra memory buffers to save the latent representation for each sample in memories.
The extra memory buffers do increase the memory cost to some extent, but still very small in compared with the whole training set.
In our implementation, we save the representation from the fc layer before the last one, which is a vector with length from 256 to 2048 for different network.
That means the cost of the representation memory is even smaller than the data memory.

\subsection{Total algorithm}

We follow A-GEM~\cite{AGEM} that unite memory of all old tasks for efficient training.
Let $\mc{M}=\cup_{k<t}\mc{M}_k$ and $\mc{F}=\cup_{k<t}\mc{F}_k$ be the united data and representation memory for old tasks.
For each step, we will sample a batch of data from the united memory.
In this way, the previous tasks will be optimized by an average gradient instead of all gradients for previous tasks, which speeds up the training.

We show the detailed process in Algorithm~\ref{alg:method} including training and evaluation procedure.
First, the storage of memory feature in \textsc{StoreMem} to be as the anchor of old task in current task training. 
Second, the gradient to be updated depends not only the gradient on old and current tasks using TAM loss, but the gradient on feature difference by ED loss.
The evaluation procedure is similar with the previous works.

\begin{table}[ht]
	\centering
	\caption{Ablation study on Split CIFAR.}
	\label{tab:abl}
	\vspace{-10px}
	\resizebox{\linewidth}{!}{
		\begin{tabular}{cc|c|cccc}
			\toprule
			$m_\text{t}$ & $m_\text{c}$ & ED & {$A_{\text{T}}(\%)$} & {$F_{\text{T}}$} & {$LCA_{10}$} & {$LTR$}  \\
			\midrule
			- & - & - & $65.44\pm1.13$ & $0.052\pm0.006$ & $\mb{0.371}\pm0.008$ & $0.377\pm0.076$  \\\midrule
			- & - & $\checkmark$ & $66.44\pm2.22$ & $0.050\pm0.009$ & $0.370\pm0.014$ & ${0.307}\pm0.066$  \\\midrule
			0.0 & 0.0 & - & $67.15\pm2.02$ & $0.053\pm0.012$ & $0.353\pm0.006$ & $0.411\pm0.097$ \\
			0.1 & 0.0 & - & $67.49\pm1.55$ & $\mb{0.049}\pm0.010$ & $0.354\pm0.005$ & $0.369\pm0.096$ \\
			0.0 & 0.01 & - & $67.45\pm1.09$ & $0.059\pm0.008$ & $0.354\pm0.005$ & $0.483\pm0.064$ \\
			0.1 & 0.01 & - & $67.68\pm1.72$ & $0.052\pm0.008$ & $0.350\pm0.007$ & $0.390\pm0.062$ \\
			0.4 & 0.01 & - & $67.28\pm0.97$ & $0.053\pm0.012$ & $0.347\pm0.006$ & $0.394\pm0.106$ \\
			0.4 & 0.05 & - & $66.68\pm1.23$ & $0.063\pm0.005$ & $0.333\pm0.005$ & $0.473\pm0.077$ \\
			0.4 & 0.1 & - & $64.97\pm1.13$ & $0.084\pm0.009$ & $0.324\pm0.008$ & $0.680\pm0.086$ \\
			\midrule
			0.1 & 0.01 & $\checkmark$ & $\mb{68.64} \pm1.35$ & $0.059 \pm0.016$ & $0.334 \pm0.008$ & $\mb{0.297} \pm0.103$ \\
			\bottomrule
	\end{tabular}}
	\vspace{-15px}
\end{table}

\begin{figure}[t]
	\begin{center}
		\includegraphics[width=\linewidth]{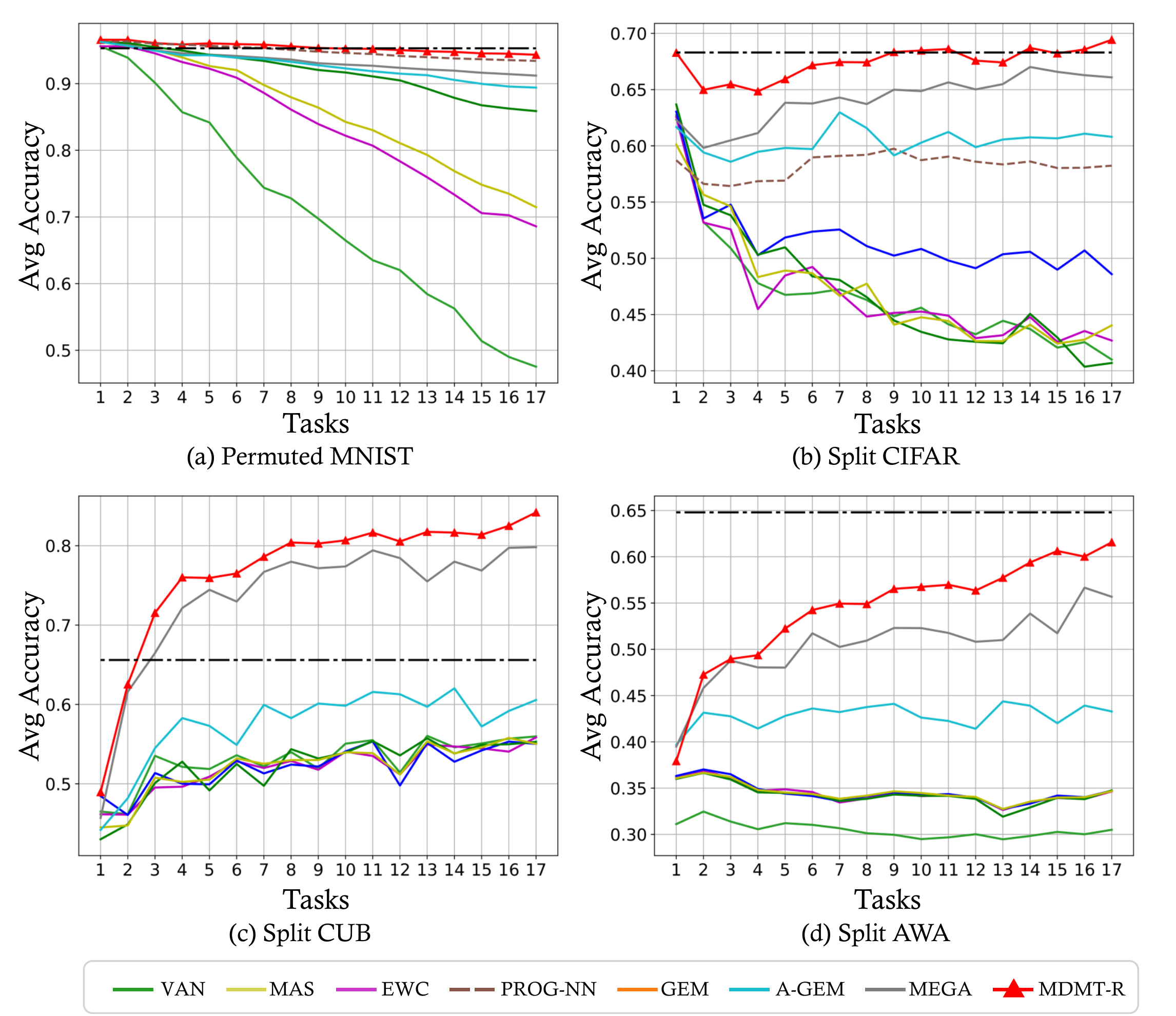}
		\caption{
			Average accuracy trend (from $A_1$ to $A_T$) on four datasets in the lifelong process.
		}
		\label{fig:acc}
	\end{center}
	\vspace{-20px}
\end{figure}

\section{Experiments}

\subsection{Experimental Settings}

We evaluate the proposed method on four image recognition datasets.
(1)\textit{Permuted MNIST.} \cite{kirkpatrick2017overcoming}: this is a variant of standard MNIST dataset of handwritten digits with 20 tasks. Each task has a fixed random permutation of the input pixels which is applied to all the images of that task.
(2)\textit{Split CIFAR.} \cite{zenke2017continual}: this dataset consists of 20 disjoint subsets of CIFAR-100 dataset \cite{krizhevsky2009learning}, where each subset is formed by randomly sampling 5 classes without replacement from the original 100 classes.
(3)\textit{Split CUB.} \cite{AGEM}: the CUB dataset~\cite{wah2011caltech} is split into 20 disjoint subsets by randomly sampling 10 classes without replacement from the original 200 classes.
(4)\textit{Split AWA.} \cite{AGEM}: this dataset consists of 20 subsets of the AWA dataset~\cite{lampert2009learning}. 
Each subset is constructed by sampling 5 classes with replacement from a total of 50 classes and the same class can appear in different subsets. 

We leverage four existing metrics to evaluate the performance and catastrophic forgetting.
(1) \textit{Average Accuracy}~($A_t\in[0,1]$) after the model has been trained continuously done till task $t\in\{1,\cdots,T\}$.
In particular, $A_T$ is the average accuracy on all the tasks after the last task has been learned.
(2) \textit{Forgetting Measure}~\cite{chaudhry2018riemannian}~($F_t\in[-1,1]$) is the 
average forgetting after the model has been trained continuously with all the mini-batches for task $t\in\{1,\cdots,T\}$.
(3) \textit{Learning Curve Area}~\cite{chaudhry2018riemannian}.
~($LCA\in[0,1]$) is the area of the convergence curve for any average $b$-shot performance after the model has been trained for all the $T$ tasks, where $b\in \left[0,\beta\right]$.
(4) \textit{Long-Term Remembering}~\cite{guo2019learning}.~($LTR\ge0$)
$LTR$ quantifies the accuracy drop on each task relative to the accuracy just right after the task has been learned.
The detailed descriptions and the formulas can be shown in the supplementary materials.


\begin{figure*}[t]
	\begin{center}
		\includegraphics[width=\linewidth]{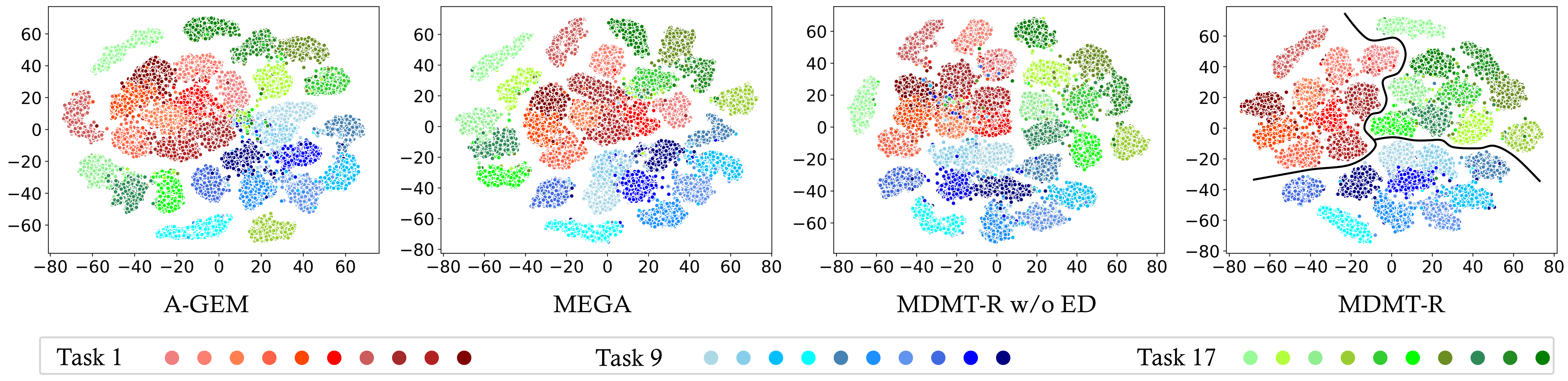}
		\caption{Final t-SNE of the features extracted from task 1, 9 and 17 on Permuted MNIST after the training on task 17.}
		\label{fig:single_class_tsne}
	\end{center}
	\vspace{-20px}
\end{figure*}

\begin{figure}[t]
	\begin{center}
		\includegraphics[width=\linewidth]{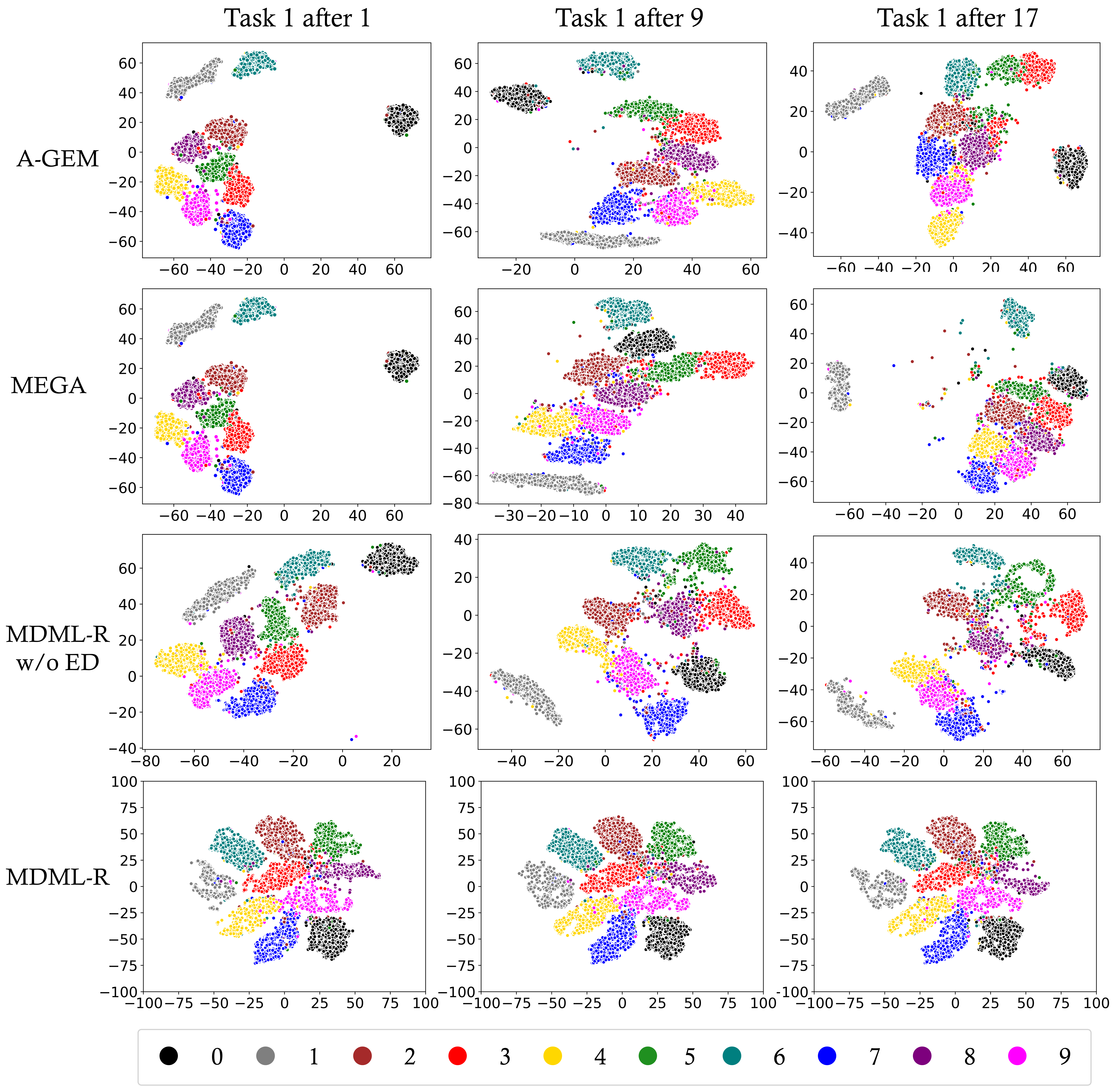}
		\caption{t-SNE of the features from task 1 on Permuted MNIST after the lifelong learning on task 1, 9 and 17.
		}
		\label{fig:mnist_tsne}
	\end{center}
	\vspace{-25px}
\end{figure}

Following the previous works~\cite{lopez2017gradient,AGEM,guo2019learning}, for Permuted MNIST we adopt a standard fully-connected network with two hidden layers, where each layer has 256 units with ReLU activation. 
For Split CIFAR we use a reduced ResNet18~\cite{he2016deep}. 
For Split CUB and Split AWA, we use a standard ResNet18.

\subsection{Comparison with the state-of-the-arts}

We compare the proposed method with the state-of-the-art methods including EWC~\cite{kirkpatrick2017overcoming}, MAS~\cite{aljundi2018memory}, RWalk~\cite{chaudhry2018riemannian}, PI~\cite{zenke2017continual}, GEM~\cite{lopez2017gradient}, MER~\cite{riemer2018learning}, ER~\cite{Chaudhry2019}  A-GEM~\cite{AGEM} and MEGA~\cite{guo2019learning}.
Specifically, EWC, MAS, RWalk and PI are regularization-based methods that prevent the important weights from changing too much.
GEM, MER, ER, AGEM and MEGA are rehearsal-based methods that rectifies the gradient guided by the stored data.
VAN is a single supervised model trained continuously on the sequence of tasks.
We also compare with the baseline that jointly trains all datasets with different classifiers together.


First, as shown in Tab.~\ref{tab:sota}, the quantitative results of the proposed method outperform other state-of-the-arts.
For $A_T$, the performances of our method show the superiority on all four datasets. 
This indicates the less forgetting on old tasks and better learning on new tasks through the lifelong training by reducing unpredictable domain shift.
$F_T$ evaluates the fine-grained batch-level forgetting on all tasks and never cares the Acc value. 
We get good $F_T$ except on Split CIFAR with slight worse ($0.04$ vs. $0.03$) than MER. 
MER has a better $F_T$ but poor $A_T$ because it adopts a complex meta learning strategy. 
For $LCA_{10}$, it evaluates the training speed on the first 10 training batches for each task, our method has the best $LCA_{10}$ only on Split CUB. 
This is because the TAM and ED losses may slow the early training to mitigate domain overlap, but the following training will be improved significantly. 
LTR focuses on long-term remembering and our method outperforms other methods on these datasets except Split CUB. 
We think this is because the dataset CUB contains similar classes of birds, which means less impact of TAM and ED losses because of similar representations.
In Fig.~\ref{fig:acc}, we show the average accuracy trends in the continual process (from $A_1$ to $A_T$), which also indicate the better performance of the MDMT-R.

In Tab.~\ref{tab:abl}, we then analyze the importance of the main components including TAM and ED loss on Split CIFAR.
The first row is the results with only vanilla softmax.
By adding ED loss, the average accuracy gets a little improvement.
By adding TAM loss, the performance obtains larger gains, and we select the best $m_\text{t}$ and $m_\text{c}$ as the hyperparameters where $m_\text{t}=0$ and $m_\text{c}=0$ means the Cross-Domain Softmax.
By adding both TAM and ED loss, we obtain a dramatic improvement in performance compared to the vanilla softmax and the state-of-the-art methods, which means the TAM and ED loss can significantly reduce the catastrophic forgetting.

\subsection{Domain shift observation}

In this section, we would like to show some observations of domain shift using t-distributed Stochastic Neighbor Embedding (t-SNE)~\cite{maaten2008visualizing} on Permuted MNIST.
First, in order to intuitively reflect the task relation of the proposed method during the training process, we visualize the final feature distribution, \ie, trained after the task 17, of task 1, 9 and 17 in Fig.~\ref{fig:single_class_tsne}.
A-GEM and MEGA cannot guarantee the task boundaries, which means generating some mix area and makes the task easy to misclassify. 
The proposed MDMT rehearsal separates each class in three task while obtain explicit task boundary, which means the proposed method is able to encourage the intra-class/task compactness and inter-class/task discrepancy.
As shown in Fig.~\ref{fig:mnist_tsne}, we also show the domain shift of task 1 after the model trained on task 1, 9 and 17, respectively.
The previous methods A-GEM and MEGA cannot reduce the domain shift at all, which makes them sustainable to forget.
The proposed MDMT rehearsal method can significantly mitigate the unpredictable
domain shift.
Without ED loss, our MDMT rehearsal still gets some unpredictable domain shift (such as task 1 after 1 and 9) because of the shrink of training data.

\section{Conclusion}
In this paper, we address catastrophic forgetting, a major drawback of state-of-the-art lifelong learning study, by considering the unpredictable domain shift of old tasks in the training sequence. 
To this end, we proposed a Multi-Domain Multi-Task rehearsal method, which effectively makes all tasks perceive each other. 
Then we proposed a Two-level Angular Margin loss to further encourage the intra-class/task compactness and inter-class/task discrepancy.
Finally, an optional Episodic Distillation loss was proposed to mitigate domain shift.
We have tested the proposed approach on four image classification benchmark datasets. 
Extensive experiments show the superiority of our approach over state-of-the-art methods. 

\section{Acknowledgment}
This work was supported by the Natural Science Foundation of China (Nos. 62072334, 61671325, 61876121, 61672376 and U1803264) and Jiangsu Provincial Key Research and Development Program (No. BE2017663). 
The authors would like to thank constructive and valuable suggestions for this paper from the experienced reviewers and AE.

{\small
	\bibliographystyle{aaai21}
	\bibliography{ref2}

\begin{thebibliography}{46}
\providecommand{\natexlab}[1]{#1}
\providecommand{\url}[1]{\texttt{#1}}
\providecommand{\urlprefix}{URL }
\expandafter\ifx\csname urlstyle\endcsname\relax
  \providecommand{\doi}[1]{doi:\discretionary{}{}{}#1}\else
  \providecommand{\doi}{doi:\discretionary{}{}{}\begingroup
  \urlstyle{rm}\Url}\fi

\bibitem[{Aljundi et~al.(2018)Aljundi, Babiloni, Elhoseiny, Rohrbach, and
  Tuytelaars}]{aljundi2018memory}
Aljundi, R.; Babiloni, F.; Elhoseiny, M.; Rohrbach, M.; and Tuytelaars, T.
  2018.
\newblock Memory aware synapses: Learning what (not) to forget.
\newblock In \emph{Proceedings of the European Conference on Computer Vision
  (ECCV)}.

\bibitem[{Chaudhry et~al.(2018{\natexlab{a}})Chaudhry, Dokania, Ajanthan, and
  Torr}]{chaudhry2018riemannian}
Chaudhry, A.; Dokania, P.~K.; Ajanthan, T.; and Torr, P.~H. 2018{\natexlab{a}}.
\newblock Riemannian walk for incremental learning: Understanding forgetting
  and intransigence.
\newblock In \emph{Proceedings of the European Conference on Computer Vision
  (ECCV)}.

\bibitem[{Chaudhry et~al.(2018{\natexlab{b}})Chaudhry, Ranzato, Rohrbach, and
  Elhoseiny}]{AGEM}
Chaudhry, A.; Ranzato, M.; Rohrbach, M.; and Elhoseiny, M. 2018{\natexlab{b}}.
\newblock Efficient Lifelong Learning with A-GEM.
\newblock In \emph{International Conference on Learning Representations}.

\bibitem[{Chaudhry et~al.(2019{\natexlab{a}})Chaudhry, Rohrbach, Elhoseiny,
  Ajanthan, Dokania, Torr, and Ranzato}]{chaudhry2019on}
Chaudhry, A.; Rohrbach, M.; Elhoseiny, M.; Ajanthan, T.; Dokania, P.~K.; Torr,
  P.~H.; and Ranzato, M. 2019{\natexlab{a}}.
\newblock On tiny episodic memories in continual learning.
\newblock \emph{arXiv preprint arXiv:1902.10486} .

\bibitem[{Chaudhry et~al.(2019{\natexlab{b}})Chaudhry, Rohrbach, Elhoseiny,
  Ajanthan, Dokania, Torr, and Ranzato}]{Chaudhry2019}
Chaudhry, A.; Rohrbach, M.; Elhoseiny, M.; Ajanthan, T.; Dokania, P.~K.; Torr,
  P. H.~S.; and Ranzato, M. 2019{\natexlab{b}}.
\newblock On Tiny Episodic Memories in Continual Learning.
\newblock \emph{arXiv preprint arXiv:1902.10486} .

\bibitem[{De~Lange et~al.(2019)De~Lange, Aljundi, Masana, Parisot, Jia,
  Leonardis, Slabaugh, and Tuytelaars}]{de2019continual}
De~Lange, M.; Aljundi, R.; Masana, M.; Parisot, S.; Jia, X.; Leonardis, A.;
  Slabaugh, G.; and Tuytelaars, T. 2019.
\newblock Continual learning: A comparative study on how to defy forgetting in
  classification tasks.
\newblock \emph{arXiv preprint arXiv:1909.08383} .

\bibitem[{Deng et~al.(2018)Deng, Wu, Wu, Hu, Lyu, and Tan}]{deng2018visual}
Deng, C.; Wu, Q.; Wu, Q.; Hu, F.; Lyu, F.; and Tan, M. 2018.
\newblock Visual grounding via accumulated attention.
\newblock In \emph{Proceedings of the IEEE conference on computer vision and
  pattern recognition}, 7746--7755.

\bibitem[{Deng et~al.(2019)Deng, Guo, Xue, and Zafeiriou}]{deng2018arcface}
Deng, J.; Guo, J.; Xue, N.; and Zafeiriou, S. 2019.
\newblock Arcface: Additive angular margin loss for deep face recognition.
\newblock In \emph{Proceedings of the IEEE Conference on Computer Vision and
  Pattern Recognition}.

\bibitem[{Dhar et~al.(2019)Dhar, Singh, Peng, Wu, and
  Chellappa}]{dhar2019learning}
Dhar, P.; Singh, R.~V.; Peng, K.-C.; Wu, Z.; and Chellappa, R. 2019.
\newblock Learning without memorizing.
\newblock In \emph{Proceedings of the IEEE Conference on Computer Vision and
  Pattern Recognition}.

\bibitem[{Fourure et~al.(2017)Fourure, Emonet, Fromont, Muselet, Neverova,
  Tr{\'{e}}meau, and Wolf}]{Fourure2017}
Fourure, D.; Emonet, R.; Fromont, {\'{E}}.; Muselet, D.; Neverova, N.;
  Tr{\'{e}}meau, A.; and Wolf, C. 2017.
\newblock Multi-task, multi-domain learning: Application to semantic
  segmentation and pose regression.
\newblock \emph{Neurocomputing} .

\bibitem[{French(1999)}]{french1999catastrophic}
French, R.~M. 1999.
\newblock Catastrophic forgetting in connectionist networks.
\newblock \emph{Trends in cognitive sciences} .

\bibitem[{Guo et~al.(2019)Guo, Liu, Yang, and Rosing}]{guo2019learning}
Guo, Y.; Liu, M.; Yang, T.; and Rosing, T. 2019.
\newblock Learning with Long-term Remembering: Following the Lead of Mixed
  Stochastic Gradient.
\newblock \emph{arXiv preprint arXiv:1909.11763} .

\bibitem[{He et~al.(2016)He, Zhang, Ren, and Sun}]{he2016deep}
He, K.; Zhang, X.; Ren, S.; and Sun, J. 2016.
\newblock Deep residual learning for image recognition.
\newblock In \emph{Proceedings of the IEEE conference on computer vision and
  pattern recognition}.

\bibitem[{Hinton, Vinyals, and Dean(2015)}]{Hinton2015}
Hinton, G.~E.; Vinyals, O.; and Dean, J. 2015.
\newblock Distilling the Knowledge in a Neural Network.
\newblock \emph{CoRR} .

\bibitem[{Hou et~al.(2018)Hou, Pan, Change~Loy, Wang, and
  Lin}]{hou2018lifelong}
Hou, S.; Pan, X.; Change~Loy, C.; Wang, Z.; and Lin, D. 2018.
\newblock Lifelong learning via progressive distillation and retrospection.
\newblock In \emph{Proceedings of the European Conference on Computer Vision
  (ECCV)}.

\bibitem[{Kirkpatrick et~al.(2017)Kirkpatrick, Pascanu, Rabinowitz, Veness,
  Desjardins, Rusu, Milan, Quan, Ramalho, Grabska-Barwinska
  et~al.}]{kirkpatrick2017overcoming}
Kirkpatrick, J.; Pascanu, R.; Rabinowitz, N.; Veness, J.; Desjardins, G.; Rusu,
  A.~A.; Milan, K.; Quan, J.; Ramalho, T.; Grabska-Barwinska, A.; et~al. 2017.
\newblock Overcoming catastrophic forgetting in neural networks.
\newblock \emph{Proceedings of the national academy of sciences} .

\bibitem[{Krizhevsky, Hinton et~al.(2009)}]{krizhevsky2009learning}
Krizhevsky, A.; Hinton, G.; et~al. 2009.
\newblock Learning multiple layers of features from tiny images .

\bibitem[{Lampert, Nickisch, and Harmeling(2009)}]{lampert2009learning}
Lampert, C.~H.; Nickisch, H.; and Harmeling, S. 2009.
\newblock Learning to detect unseen object classes by between-class attribute
  transfer.
\newblock In \emph{2009 IEEE Conference on Computer Vision and Pattern
  Recognition}.

\bibitem[{Lesort et~al.(2019)Lesort, Gepperth, Stoian, and
  Filliat}]{Lesort2018}
Lesort, T.; Gepperth, A.; Stoian, A.; and Filliat, D. 2019.
\newblock Marginal replay vs conditional replay for continual learning.
\newblock In \emph{International Conference on Artificial Neural Networks}.

\bibitem[{Li and Hoiem(2016)}]{li2016learning}
Li, Z.; and Hoiem, D. 2016.
\newblock Learning Without Forgetting.
\newblock In \emph{European Conference on Computer Vision}.

\bibitem[{Lin et~al.(2019)Lin, Zhen, Li, Zhang, and Kwong}]{lin2019pareto}
Lin, X.; Zhen, H.-L.; Li, Z.; Zhang, Q.-F.; and Kwong, S. 2019.
\newblock Pareto Multi-Task Learning.
\newblock In \emph{Advances in Neural Information Processing Systems}.

\bibitem[{Liu et~al.(2017)Liu, Wen, Yu, Li, Raj, and
  Song}]{DBLP:conf/cvpr/LiuWYLRS17}
Liu, W.; Wen, Y.; Yu, Z.; Li, M.; Raj, B.; and Song, L. 2017.
\newblock SphereFace: Deep Hypersphere Embedding for Face Recognition.
\newblock In \emph{2017 {IEEE} Conference on Computer Vision and Pattern
  Recognition, {CVPR} 2017, Honolulu, HI, USA, July 21-26, 2017}.

\bibitem[{Liu et~al.(2016)Liu, Wen, Yu, and Yang}]{liu2016large-margin}
Liu, W.; Wen, Y.; Yu, Z.; and Yang, M. 2016.
\newblock Large-margin softmax loss for convolutional neural networks.
\newblock In \emph{ICML}.

\bibitem[{Lopez-Paz and Ranzato(2017)}]{lopez2017gradient}
Lopez-Paz, D.; and Ranzato, M. 2017.
\newblock Gradient episodic memory for continual learning.
\newblock In \emph{Advances in neural information processing systems}.

\bibitem[{Lyu, Feng, and Wang(2020)}]{lyu2020vtgraphnet}
Lyu, F.; Feng, W.; and Wang, S. 2020.
\newblock vtGraphNet: Learning weakly-supervised scene graph for complex visual
  grounding.
\newblock \emph{Neurocomputing} 51--60.

\bibitem[{Lyu et~al.(2019)Lyu, Wu, Hu, Wu, and Tan}]{lyu2019attend}
Lyu, F.; Wu, Q.; Hu, F.; Wu, Q.; and Tan, M. 2019.
\newblock Attend and imagine: Multi-label image classification with visual
  attention and recurrent neural networks.
\newblock \emph{IEEE Transactions on Multimedia} 1971--1981.

\bibitem[{Maaten and Hinton(2008)}]{maaten2008visualizing}
Maaten, L. v.~d.; and Hinton, G. 2008.
\newblock Visualizing data using t-SNE.
\newblock \emph{Journal of machine learning research} .

\bibitem[{Mallya, Davis, and Lazebnik(2018)}]{mallya2018piggyback}
Mallya, A.; Davis, D.; and Lazebnik, S. 2018.
\newblock Piggyback: Adapting a single network to multiple tasks by learning to
  mask weights.
\newblock In \emph{Proceedings of the European Conference on Computer Vision
  (ECCV)}.

\bibitem[{Nam and Han(2016)}]{Nam2015}
Nam, H.; and Han, B. 2016.
\newblock Learning multi-domain convolutional neural networks for visual
  tracking.
\newblock In \emph{Proceedings of the IEEE conference on computer vision and
  pattern recognition}.

\bibitem[{Pellegrini et~al.(2019)Pellegrini, Graffieti, Lomonaco, and
  Maltoni}]{Pellegrini2019}
Pellegrini, L.; Graffieti, G.; Lomonaco, V.; and Maltoni, D. 2019.
\newblock Latent replay for real-time continual learning.
\newblock \emph{arXiv preprint arXiv:1912.01100} .

\bibitem[{Peng and Dredze(2016)}]{Peng2016}
Peng, N.; and Dredze, M. 2016.
\newblock Multi-task Multi-domain Representation Learning for Sequence Tagging.
\newblock \emph{CoRR} .

\bibitem[{Rebuffi, Kolesnikov, and Lampert(2016)}]{Rebuffi2016}
Rebuffi, S.; Kolesnikov, A.; and Lampert, C.~H. 2016.
\newblock iCaRL: Incremental Classifier and Representation Learning.
\newblock \emph{CoRR} .

\bibitem[{Riemer et~al.(2018)Riemer, Cases, Ajemian, Liu, Rish, Tu, and
  Tesauro}]{riemer2018learning}
Riemer, M.; Cases, I.; Ajemian, R.; Liu, M.; Rish, I.; Tu, Y.; and Tesauro, G.
  2018.
\newblock Learning to learn without forgetting by maximizing transfer and
  minimizing interference.
\newblock \emph{arXiv preprint arXiv:1810.11910} .

\bibitem[{Ring(1997)}]{ring1997child}
Ring, M.~B. 1997.
\newblock CHILD: A first step towards continual learning.
\newblock \emph{Machine Learning} .

\bibitem[{Rolnick et~al.(2019)Rolnick, Ahuja, Schwarz, Lillicrap, and
  Wayne}]{rolnick2018experience}
Rolnick, D.; Ahuja, A.; Schwarz, J.; Lillicrap, T.; and Wayne, G. 2019.
\newblock Experience replay for continual learning.
\newblock In \emph{Advances in Neural Information Processing Systems}.

\bibitem[{Sener and Koltun(2018)}]{sener2018multi}
Sener, O.; and Koltun, V. 2018.
\newblock Multi-task learning as multi-objective optimization.
\newblock In \emph{Advances in Neural Information Processing Systems}.

\bibitem[{Shen et~al.(2020)Shen, Zhang, Chen, and Deng}]{Shen2020}
Shen, G.; Zhang, S.; Chen, X.; and Deng, Z.-H. 2020.
\newblock Generative Feature Replay with Orthogonal Weight Modification for
  Continual Learning.
\newblock \emph{arXiv preprint arXiv:2005.03490} .

\bibitem[{Tang and Jia(2020)}]{Tang2020}
Tang, H.; and Jia, K. 2020.
\newblock Discriminative Adversarial Domain Adaptation.
\newblock In \emph{The Thirty-Fourth {AAAI} Conference on Artificial
  Intelligence, {AAAI} 2020, The Thirty-Second Innovative Applications of
  Artificial Intelligence Conference, {IAAI} 2020, The Tenth {AAAI} Symposium
  on Educational Advances in Artificial Intelligence, {EAAI} 2020, New York,
  NY, USA, February 7-12, 2020}.

\bibitem[{Thrun(1998)}]{thrun1998lifelong}
Thrun, S. 1998.
\newblock Lifelong learning algorithms.
\newblock In \emph{Learning to learn}. Springer.

\bibitem[{van~de Ven and Tolias(2018)}]{Ven2018}
van~de Ven, G.~M.; and Tolias, A.~S. 2018.
\newblock Generative replay with feedback connections as a general strategy for
  continual learning.
\newblock \emph{arXiv preprint arXiv:1809.10635} .

\bibitem[{Wah et~al.(2011)Wah, Branson, Welinder, Perona, and
  Belongie}]{wah2011caltech}
Wah, C.; Branson, S.; Welinder, P.; Perona, P.; and Belongie, S. 2011.
\newblock The caltech-ucsd birds-200-2011 dataset .

\bibitem[{Wang et~al.(2018{\natexlab{a}})Wang, Cheng, Liu, and
  Liu}]{DBLP:journals/spl/WangCLL18}
Wang, F.; Cheng, J.; Liu, W.; and Liu, H. 2018{\natexlab{a}}.
\newblock Additive Margin Softmax for Face Verification.
\newblock \emph{{IEEE} Signal Process. Lett.} .

\bibitem[{Wang et~al.(2018{\natexlab{b}})Wang, Wang, Zhou, Ji, Gong, Zhou, Li,
  and Liu}]{DBLP:conf/cvpr/WangWZJGZL018}
Wang, H.; Wang, Y.; Zhou, Z.; Ji, X.; Gong, D.; Zhou, J.; Li, Z.; and Liu, W.
  2018{\natexlab{b}}.
\newblock CosFace: Large Margin Cosine Loss for Deep Face Recognition.
\newblock In \emph{2018 {IEEE} Conference on Computer Vision and Pattern
  Recognition, {CVPR} 2018, Salt Lake City, UT, USA, June 18-22, 2018}.

\bibitem[{Yang and Hospedales(2014)}]{Yang2014}
Yang, Y.; and Hospedales, T.~M. 2014.
\newblock A Unified Perspective on Multi-Domain and Multi-Task Learning.
\newblock In \emph{International Conference on Learning Representations}.

\bibitem[{Yoon et~al.(2017)Yoon, Yang, Lee, and Hwang}]{yoon2017lifelong}
Yoon, J.; Yang, E.; Lee, J.; and Hwang, S.~J. 2017.
\newblock Lifelong learning with dynamically expandable networks.
\newblock \emph{arXiv preprint arXiv:1708.01547} .

\bibitem[{Zenke, Poole, and Ganguli(2017)}]{zenke2017continual}
Zenke, F.; Poole, B.; and Ganguli, S. 2017.
\newblock Continual learning through synaptic intelligence.
\newblock \emph{Proceedings of machine learning research} .

\end{thebibliography}
}

\newpage
\section{Appendix}
\subsection{Evaluation Metrics}
\noindent
\textbf{Average Accuracy.}~($A_t\in[0,1]$)
Average accuracy after the model has been trained continually
with all the mini-batches up till task $t\in\{1,\cdots,T\}$ is defined as
\begin{equation}
	A_{t}=\frac{1}{t} \sum_{j=1}^{t} a_{t, B_{t}, j},
\end{equation}
where $a_{k, B_{t}, j}$ is the mean accuracy of task $j$ on $B_{t}$ mini-batches.
In particular, $A_T$ is the average accuracy on all the tasks after the last task has been learned.

\noindent
\textbf{Forgetting Measure}~\cite{chaudhry2018riemannian}.~($F_t\in[-1,1]$) 
Average forgetting after the model has been trained continually with all the mini-batches up till task $t\in\{1,\cdots,T\}$ is defined as
\begin{equation}
	F_{t}=\frac{1}{t-1} \sum_{j=1}^{t-1} f_{j}^{t}
\end{equation}
where $f_{j}^{t}$ is the forgetting on task $j$ after the model is trained with all the mini-batches up till task $t$ and computed as
\begin{equation}
	f_{j}^{t}=\max _{l \in\{1, \cdots, k-1\}} a_{l, B_{l}, j}-a_{t, B_{k}, j}
\end{equation}

\noindent
\textbf{Learning Curve Area}~\cite{chaudhry2018riemannian}.
~($LCA\in[0,1]$) 
$LCA_\beta$ is the area of the convergence curve for any average $b$-shot performance after the model has been trained for all the $T$ tasks, where $b\in \left[0,\beta\right]$.
\begin{equation}
	\mathrm{LCA}_{\beta}=\frac{1}{\beta+1} \int_{0}^{\beta} Z_{b} d b=\frac{1}{\beta+1} \sum_{b=0}^{\beta} Z_{b},
\end{equation}
where
\begin{equation}
	Z_{b}=\frac{1}{T} \sum_{t=1}^{T} a_{t, b, t}.
\end{equation}
Intuitively, LCA measures the learning speed of different lifelong
learning algorithms. 
A higher value of LCA indicates that the model learns quickly.

\noindent
\textbf{Long-Term Remembering}~\cite{guo2019learning}.~($LTR\ge0$)
$LTR$ quantifies the accuracy drop on each task relative to the accuracy just right after the task has been learned, which is defined as
\begin{equation}
	\mathrm{LTR}=-\frac{1}{T-1} \sum_{j=1}^{T-1}(T-j) \min \left\{0, a_{T, B_{T}, j}-a_{j, B_{j}, j}\right\}
\end{equation}
After the model is trained on all the $T$ tasks, LTR quantifies the accuracy drop on task $D_j$ relative to $a_{j, B_{j}, j}$.

\subsection{Implementation details}

\noindent
\textbf{Seed initialization.} 
We implement our method with Tensorflow. 
The results are the average of 5 runs, where the seeds of Numpy and Tensorflow are both from 1234 to 1238 for all compared methods.

\noindent
\textbf{Selected classes.} 
For Split CUB and AWA, AGEM has provided fixed classes selections. 
For Perm MNIST, each task has a fixed random permutation by shuffling the pixels in images, the seeds are the same with the training. 
Split CIFAR has 20 disjoint subsets by extracted each 5-classes in sequence from class 1 to 100.
Train/test splits: For Perm MNIST and Split CIFAR, the data splits are the same as vanilla MNIST and CIFAR. For Split CUB and AWA, AGEM has provided fixed train/test list.

\noindent
\textbf{Hyper-Parameters selection.}
We report the hyper-parameters considered for different experiments of $m_\text{t}$, $m_\text{c}$ and $s$. 
The searching spaces for the hyper parameters are: 
$m_\text{t}$: [0.01, 0.05, 0.1 (Permuted MNIST, Split CIFAR), 0.2 (Split AWA), 0.3, 0.4 (Split CUB), 0.5]; 
$m_\text{c}$: [0.01 (Permuted MNIST, Split CIFAR), 0.02 (Split AWA), 0.03, 0.04, 0.05 (Split CUB), 0.06, …, 0.1];
$s$: [12, 16, 20 (Split CUB), 24 (Split CIFAR), 28, 32 (Permuted MNIST, Split AWA), 36, …, 64].
Note that we do not conduct exhaustive hyper-parameter search on learning rate and regularization factors and reuse the same as A-GEM~\cite{AGEM}.
The learning rates for each dataset are: Perm MNIST: $0.1$, Split CIFAR and CUB: $0.03$, Split AWA: $0.01$.

%
\begin{table*}[t]
	\centering
	\caption{Computational cost and memory complexity. The memory cost consists of several parameters: (1) the number of task $T$; (2) the total number of parameters $P$; (3) the size of the mini-batch $B$; (4) the total size of the network hidden state $H$ (assuming all methods use the same architecture); (5) the size of the episodic memory $M$ per task; (6) the size of the $l$-th layer hidden state $\tilde{H}$ \wrt the latent representation.}
	\label{tab:cost}
	\resizebox{0.75\linewidth}{!}{
		\begin{tabular}{l|cccc|cc}
			\toprule
			\multirow{2}{*}{Method}  & \multicolumn{4}{c|}{\textbf{Inference time (ms)}} & \multicolumn{2}{c}{\textbf{Memory}}  \\
			\cline{2-7}
			& MNIST& CIFAR & CUB & AWA & Training & Testing \\
			\midrule
			VAN 	& - & - & - & - & $P+B*H$ & $P+B*H$ \\
			EWC 	& - & - & - & - & $4*P+B*H$ & $4*P+B*H$ \\
			GEM 	& - & - & - & - & $P*T+(B+M)*H$ & $P+B*H$ \\
			A-GEM 	& 32.58 & 28.74 & 81.83 & 97.47 & $2*P+(B+M)*H$ & $P+B*H$ \\
			MEGA 	& 37.63 & 26.38 & 84.96 & 109.67 & $2*P+(B+M)*H$ & $P+B*H$ \\
			MDMT-R 	& 35.16 & 30.09 & 90.47 & 97.51 & $2*P+(B+M)*H+M*H_l$ & $P+B*H$ \\
			\bottomrule
	\end{tabular}}
\end{table*}

\subsection{Evolution of LCA}

\begin{figure}[t]
	\begin{center}
		\includegraphics[width=\linewidth]{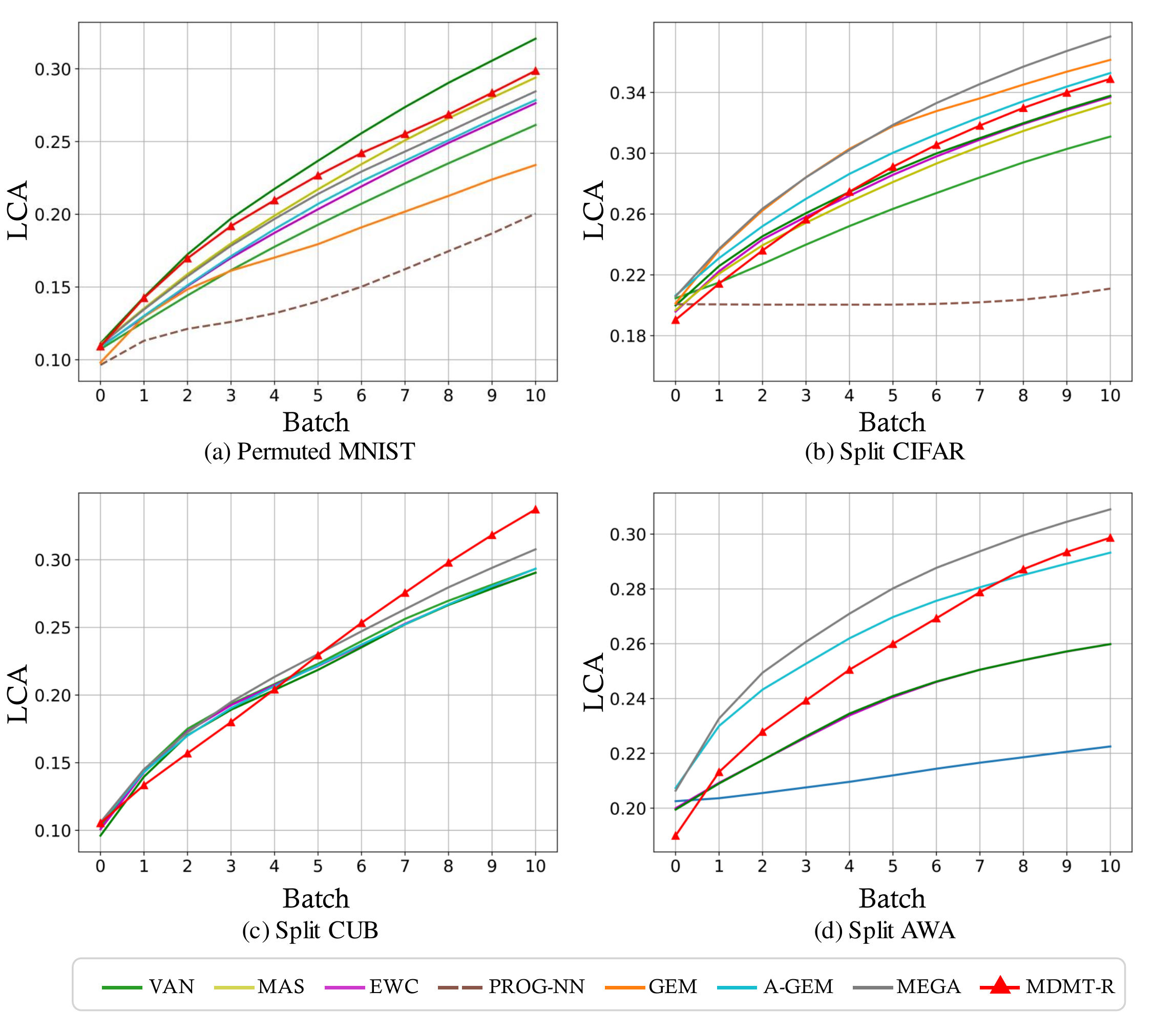}
		\caption{Evolution of LCA during the first ten mini-batches.}
		\label{fig:lca}
	\end{center}
\end{figure}

We show the evolution of $LCA$ during the first ten mini-batches on the four datasets, where $LCA_k$ represents the learning speed of new task in the first $k$ mini-batches.
As shown in Fig.~\ref{fig:lca}, the proposed MDMT rehearsal based lifelong learning outperforms other methods on Split CUB, however a slight worse than MEGA.
In our option, this is because the ED loss tries to make the model not fall into catastrophic forgetting at the training beginning.
Thus, the proposed MDMT rehearsal based lifelong learning do not achieve fast improvements at the beginning of each task.

\subsection{Computational cost and memory complexity}

We show the computational cost and memory complexity in Table~\ref{tab:cost} by evaluating the training time on single RTX 2080Ti GPU card.
Easy to see, the inference time of the proposed method is comparable with A-GEM and MEGA on four datasets.
We can also see the memory cost of training and testing procedure in the right of Table~\ref{tab:cost}.
Compared to the previous baselines based on EpsMem, the cost of the proposed method has a slight increase $M*\tilde{H}$, which arises from the saving of latent representation of old tasks.
However, such an increase is very small compared to the total training cost because in most situations the latent representation (a vector) has smaller size than the image (a 3-channel tensor).

\subsection{Raw Record of Learning Process}

In this section, we report the raw record of learning process for the proposed method, MEGA and A-GEM.
The entry $(i,j)$ of the matrix for each method is the test accuracy of the $j$-th task after the model is trained on the $i$-th task.

\subsubsection{PERMUTED MNIST}

\paragraph{MDMT-R:}

~

\begin{spacing}{.5}
	\noindent\resizebox{\linewidth}{!}{0.9658 0.1207 0.1267 0.0846 0.1393 0.0835 0.1005 0.0883 0.1218 0.1004 0.0952 0.1267 0.1263 0.1041 0.0929 0.0910 0.0995}
	\noindent\resizebox{\linewidth}{!}{0.9638 0.9676 0.1059 0.0995 0.1204 0.0740 0.0849 0.0955 0.1126 0.1292 0.1146 0.1180 0.1302 0.0945 0.0867 0.1021 0.0981}
	\noindent\resizebox{\linewidth}{!}{0.9574 0.9638 0.9612 0.0842 0.0935 0.0925 0.0793 0.1035 0.1002 0.1226 0.1073 0.1043 0.1086 0.1024 0.0962 0.0706 0.0822}
	\noindent\resizebox{\linewidth}{!}{0.9548 0.9596 0.9652 0.9554 0.0954 0.0790 0.1026 0.1120 0.0935 0.1198 0.0961 0.1186 0.1247 0.0833 0.1099 0.0753 0.0821}
	\noindent\resizebox{\linewidth}{!}{0.9498 0.9578 0.9607 0.9660 0.9679 0.0854 0.0889 0.1347 0.1134 0.1258 0.1014 0.1410 0.1396 0.1091 0.0994 0.0822 0.1033}
	\noindent\resizebox{\linewidth}{!}{0.9477 0.9544 0.9595 0.9627 0.9655 0.9662 0.0998 0.1316 0.1121 0.1262 0.1203 0.1498 0.1303 0.0968 0.1044 0.0817 0.0903}
	\noindent\resizebox{\linewidth}{!}{0.9451 0.9532 0.9568 0.9607 0.9630 0.9636 0.9669 0.1310 0.1004 0.1341 0.1004 0.1286 0.1438 0.1127 0.1124 0.0834 0.0990}
	\noindent\resizebox{\linewidth}{!}{0.9415 0.9503 0.9550 0.9559 0.9610 0.9588 0.9644 0.9612 0.0977 0.1129 0.0831 0.1223 0.1353 0.1027 0.1154 0.0748 0.0992}
	\noindent\resizebox{\linewidth}{!}{0.9359 0.9466 0.9483 0.9543 0.9574 0.9557 0.9600 0.9609 0.9625 0.1141 0.1028 0.1318 0.1331 0.1005 0.1128 0.0909 0.1003}
	\noindent\resizebox{\linewidth}{!}{0.9326 0.9424 0.9466 0.9529 0.9536 0.9563 0.9572 0.9590 0.9632 0.9621 0.0958 0.1082 0.1196 0.1095 0.1116 0.0762 0.1005}
	\noindent\resizebox{\linewidth}{!}{0.9313 0.9426 0.9465 0.9499 0.9512 0.9528 0.9556 0.9557 0.9598 0.9627 0.9660 0.1218 0.1369 0.1405 0.1088 0.0821 0.1195}
	\noindent\resizebox{\linewidth}{!}{0.9273 0.9397 0.9438 0.9450 0.9468 0.9513 0.9506 0.9540 0.9571 0.9605 0.9640 0.9630 0.1645 0.1249 0.0974 0.0684 0.1202}
	\noindent\resizebox{\linewidth}{!}{0.9224 0.9363 0.9376 0.9446 0.9469 0.9466 0.9498 0.9496 0.9541 0.9588 0.9600 0.9619 0.9630 0.1373 0.1011 0.0633 0.1328}
	\noindent\resizebox{\linewidth}{!}{0.9178 0.9326 0.9377 0.9418 0.9419 0.9453 0.9481 0.9484 0.9518 0.9562 0.9586 0.9594 0.9618 0.9649 0.1074 0.0798 0.1247}
	\noindent\resizebox{\linewidth}{!}{0.9167 0.9296 0.9341 0.9349 0.9356 0.9439 0.9417 0.9460 0.9495 0.9535 0.9553 0.9558 0.9596 0.9623 0.9629 0.0775 0.1267}
	\noindent\resizebox{\linewidth}{!}{0.9132 0.9269 0.9314 0.9338 0.9346 0.9418 0.9414 0.9443 0.9478 0.9511 0.9533 0.9544 0.9559 0.9597 0.9628 0.9663 0.1293}
	\noindent\resizebox{\linewidth}{!}{0.9111 0.9221 0.9277 0.9303 0.9341 0.9387 0.9398 0.9409 0.9451 0.9491 0.9495 0.9520 0.9555 0.9566 0.9595 0.9625 0.9621}

\end{spacing}

\paragraph{MEGA:}

~

\begin{spacing}{.5}
	\noindent
	\resizebox{\linewidth}{!}{0.9613 0.1091 0.1229 0.0832 0.1374 0.0708 0.0907 0.1017 0.1165 0.1286 0.0979 0.1182 0.1188 0.0886 0.0968 0.0854 0.0928}
	
	\noindent\resizebox{\linewidth}{!}{0.9535 0.9645 0.0895 0.0997 0.1191 0.0685 0.0803 0.1022 0.1165 0.1472 0.1054 0.1112 0.1264 0.1027 0.0872 0.0979 0.0993}
	
	\noindent\resizebox{\linewidth}{!}{0.9391 0.9556 0.9596 0.0996 0.1020 0.0900 0.0807 0.1083 0.0959 0.1400 0.1001 0.1012 0.1096 0.1085 0.0977 0.0716 0.0768}
	
	\noindent\resizebox{\linewidth}{!}{0.9295 0.9473 0.9527 0.9477 0.1113 0.0725 0.0856 0.1033 0.0884 0.1209 0.0847 0.1149 0.1285 0.0939 0.1193 0.0867 0.0824}
	
	\noindent\resizebox{\linewidth}{!}{0.9206 0.9405 0.9437 0.9569 0.9611 0.0785 0.0884 0.1113 0.0926 0.1189 0.0936 0.1337 0.1544 0.1154 0.1282 0.1010 0.0994}
	
	\noindent\resizebox{\linewidth}{!}{0.9119 0.9347 0.9378 0.9481 0.9547 0.9594 0.0916 0.1200 0.1013 0.1042 0.0908 0.1380 0.1415 0.1199 0.1210 0.0908 0.0847}
	
	\noindent\resizebox{\linewidth}{!}{0.9083 0.9261 0.9348 0.9419 0.9478 0.9556 0.9575 0.1092 0.1040 0.1083 0.0863 0.1202 0.1286 0.1177 0.1250 0.0801 0.0931}
	
	\noindent\resizebox{\linewidth}{!}{0.9100 0.9192 0.9291 0.9332 0.9419 0.9462 0.9527 0.9598 0.1152 0.1132 0.0945 0.1054 0.1248 0.1228 0.1187 0.0945 0.0934}
	
	\noindent\resizebox{\linewidth}{!}{0.9022 0.9133 0.9215 0.9271 0.9344 0.9381 0.9430 0.9476 0.9551 0.1187 0.1162 0.1119 0.1364 0.1249 0.1108 0.1012 0.1059}
	
	\noindent\resizebox{\linewidth}{!}{0.8974 0.9074 0.9147 0.9242 0.9289 0.9348 0.9367 0.9404 0.9519 0.9571 0.1102 0.1032 0.1536 0.1261 0.1122 0.1036 0.1093}
	
	\noindent\resizebox{\linewidth}{!}{0.8957 0.9042 0.9146 0.9193 0.9229 0.9313 0.9321 0.9318 0.9409 0.9545 0.9591 0.1144 0.1368 0.1373 0.1143 0.1092 0.1169}
	
	\noindent\resizebox{\linewidth}{!}{0.8863 0.8981 0.9056 0.9127 0.9148 0.9220 0.9270 0.9249 0.9335 0.9444 0.9528 0.9564 0.1517 0.1103 0.1051 0.1002 0.1390}
	
	\noindent\resizebox{\linewidth}{!}{0.8840 0.8992 0.9054 0.9085 0.9149 0.9179 0.9238 0.9198 0.9304 0.9419 0.9441 0.9523 0.9570 0.1292 0.1083 0.0926 0.1301}
	
	\noindent\resizebox{\linewidth}{!}{0.8808 0.8901 0.8994 0.8986 0.9084 0.9113 0.9168 0.9185 0.9239 0.9360 0.9382 0.9453 0.9522 0.9589 0.1017 0.0941 0.1277}
	
	\noindent\resizebox{\linewidth}{!}{0.8770 0.8850 0.8957 0.8926 0.9012 0.9101 0.9090 0.9132 0.9188 0.9301 0.9358 0.9368 0.9430 0.9508 0.9521 0.0946 0.1334}
	
	\noindent\resizebox{\linewidth}{!}{0.8752 0.8806 0.8911 0.8854 0.8965 0.9070 0.9062 0.9059 0.9145 0.9265 0.9286 0.9338 0.9374 0.9434 0.9462 0.9601 0.1291}
	
	\noindent\resizebox{\linewidth}{!}{0.8732 0.8765 0.8824 0.8809 0.8945 0.9024 0.9016 0.9007 0.9088 0.9202 0.9228 0.9276 0.9279 0.9376 0.9408 0.9521 0.9556}
	
\end{spacing}

\paragraph{A-GEM:}

~

\begin{spacing}{.5}
	\noindent
	\resizebox{\linewidth}{!}{0.9613 0.1091 0.1229 0.0832 0.1374 0.0708 0.0907 0.1017 0.1165 0.1286 0.0979 0.1182 0.1188 0.0886 0.0968 0.0854 0.0928}
	
	\noindent\resizebox{\linewidth}{!}{0.9509 0.9645 0.0956 0.0991 0.1304 0.0696 0.0840 0.1033 0.1219 0.1454 0.1064 0.1133 0.1314 0.1043 0.0883 0.0979 0.0973}
	
	\noindent\resizebox{\linewidth}{!}{0.9410 0.9545 0.9615 0.0995 0.0964 0.0921 0.0710 0.1126 0.1176 0.1402 0.1112 0.1026 0.1185 0.1204 0.1077 0.0779 0.0761}
	
	\noindent\resizebox{\linewidth}{!}{0.9299 0.9450 0.9540 0.9546 0.1046 0.0788 0.0959 0.1033 0.1096 0.1266 0.1015 0.1152 0.1476 0.0885 0.1375 0.0984 0.0831}
	
	\noindent\resizebox{\linewidth}{!}{0.9151 0.9361 0.9425 0.9551 0.9588 0.0803 0.0809 0.1143 0.1063 0.1227 0.1066 0.1253 0.1436 0.1154 0.1131 0.1079 0.0915}
	
	\noindent\resizebox{\linewidth}{!}{0.9068 0.9312 0.9401 0.9450 0.9566 0.9590 0.0892 0.1189 0.1285 0.1086 0.1007 0.1433 0.1279 0.1179 0.1097 0.0892 0.0865}
	
	\noindent\resizebox{\linewidth}{!}{0.9015 0.9228 0.9339 0.9385 0.9473 0.9548 0.9586 0.1063 0.1073 0.1102 0.1048 0.1164 0.1291 0.1284 0.1341 0.0854 0.1024}
	
	\noindent\resizebox{\linewidth}{!}{0.8980 0.9155 0.9248 0.9280 0.9356 0.9403 0.9539 0.9580 0.1015 0.1231 0.1129 0.1125 0.1267 0.1133 0.1220 0.0921 0.0985}
	
	\noindent\resizebox{\linewidth}{!}{0.8952 0.9055 0.9201 0.9182 0.9273 0.9310 0.9447 0.9435 0.9512 0.1098 0.1374 0.1166 0.1264 0.1064 0.1183 0.0986 0.1048}
	
	\noindent\resizebox{\linewidth}{!}{0.8846 0.8996 0.9083 0.9154 0.9189 0.9267 0.9363 0.9339 0.9513 0.9558 0.1243 0.1095 0.1179 0.1137 0.1126 0.0945 0.0979}
	
	\noindent\resizebox{\linewidth}{!}{0.8764 0.8977 0.9011 0.9073 0.9086 0.9167 0.9292 0.9274 0.9386 0.9481 0.9631 0.1116 0.1099 0.1417 0.1100 0.0975 0.1166}
	
	\noindent\resizebox{\linewidth}{!}{0.8710 0.8882 0.8937 0.8922 0.9043 0.9077 0.9151 0.9174 0.9279 0.9324 0.9518 0.9572 0.1346 0.1240 0.0964 0.0930 0.1283}
	
	\noindent\resizebox{\linewidth}{!}{0.8625 0.8822 0.8847 0.8855 0.9013 0.8990 0.9093 0.9088 0.9189 0.9295 0.9411 0.9458 0.9533 0.1309 0.1059 0.0987 0.1139}
	
	\noindent\resizebox{\linewidth}{!}{0.8581 0.8784 0.8774 0.8817 0.8954 0.8938 0.8986 0.9003 0.9082 0.9225 0.9307 0.9350 0.9435 0.9603 0.1048 0.1023 0.1048}
	
	\noindent\resizebox{\linewidth}{!}{0.8492 0.8674 0.8732 0.8697 0.8828 0.8826 0.8930 0.8898 0.8962 0.9098 0.9184 0.9267 0.9290 0.9425 0.9542 0.1012 0.1070}
	
	\noindent\resizebox{\linewidth}{!}{0.8322 0.8700 0.8644 0.8493 0.8765 0.8787 0.8904 0.8848 0.8883 0.8979 0.9110 0.9158 0.9177 0.9299 0.9403 0.9609 0.1179}
	
	\noindent\resizebox{\linewidth}{!}{0.8438 0.8603 0.8555 0.8488 0.8864 0.8785 0.8798 0.8702 0.8916 0.8968 0.9076 0.9094 0.9092 0.9228 0.9228 0.9463 0.9551}
	
\end{spacing}

\subsubsection{SPLIT CIFAR}

\paragraph{MDMT-R:}

~

\begin{spacing}{.5}
	\noindent\resizebox{\linewidth}{!}{0.6828 0.0000 0.0000 0.0000 0.0000 0.0000 0.0000 0.0000 0.0000 0.0000 0.0000 0.0000 0.0000 0.0000 0.0000 0.0000 0.0000}
	\noindent\resizebox{\linewidth}{!}{0.6704 0.6292 0.0000 0.0000 0.0000 0.0000 0.0000 0.0000 0.0000 0.0000 0.0000 0.0000 0.0000 0.0000 0.0000 0.0000 0.0000}
	\noindent\resizebox{\linewidth}{!}{0.6912 0.6136 0.6596 0.0000 0.0000 0.0000 0.0000 0.0000 0.0000 0.0000 0.0000 0.0000 0.0000 0.0000 0.0000 0.0000 0.0000}
	\noindent\resizebox{\linewidth}{!}{0.6732 0.5828 0.6316 0.7064 0.0000 0.0000 0.0000 0.0000 0.0000 0.0000 0.0000 0.0000 0.0000 0.0000 0.0000 0.0000 0.0000}
	\noindent\resizebox{\linewidth}{!}{0.6604 0.5944 0.6288 0.7012 0.7120 0.0000 0.0000 0.0000 0.0000 0.0000 0.0000 0.0000 0.0000 0.0000 0.0000 0.0000 0.0000}
	\noindent\resizebox{\linewidth}{!}{0.6736 0.6140 0.5988 0.6792 0.7064 0.7584 0.0000 0.0000 0.0000 0.0000 0.0000 0.0000 0.0000 0.0000 0.0000 0.0000 0.0000}
	\noindent\resizebox{\linewidth}{!}{0.6680 0.5944 0.6068 0.6752 0.6976 0.7800 0.6996 0.0000 0.0000 0.0000 0.0000 0.0000 0.0000 0.0000 0.0000 0.0000 0.0000}
	\noindent\resizebox{\linewidth}{!}{0.6824 0.6036 0.6096 0.6780 0.6780 0.7236 0.6952 0.7236 0.0000 0.0000 0.0000 0.0000 0.0000 0.0000 0.0000 0.0000 0.0000}
	\noindent\resizebox{\linewidth}{!}{0.6716 0.6028 0.6084 0.6812 0.6840 0.7396 0.6772 0.7120 0.7744 0.0000 0.0000 0.0000 0.0000 0.0000 0.0000 0.0000 0.0000}
	\noindent\resizebox{\linewidth}{!}{0.6872 0.6136 0.6140 0.6848 0.6864 0.7432 0.6728 0.6916 0.7492 0.7052 0.0000 0.0000 0.0000 0.0000 0.0000 0.0000 0.0000}
	\noindent\resizebox{\linewidth}{!}{0.6604 0.6040 0.6104 0.6908 0.6788 0.7368 0.6764 0.6980 0.7628 0.6700 0.7592 0.0000 0.0000 0.0000 0.0000 0.0000 0.0000}
	\noindent\resizebox{\linewidth}{!}{0.6548 0.5888 0.6136 0.6784 0.6832 0.7136 0.6836 0.6744 0.7440 0.6472 0.7460 0.6820 0.0000 0.0000 0.0000 0.0000 0.0000}
	\noindent\resizebox{\linewidth}{!}{0.6484 0.5964 0.6100 0.6640 0.6744 0.7224 0.6540 0.6804 0.7224 0.6504 0.7264 0.6728 0.7424 0.0000 0.0000 0.0000 0.0000}
	\noindent\resizebox{\linewidth}{!}{0.6628 0.5936 0.6152 0.7008 0.6940 0.7272 0.6624 0.6788 0.7372 0.6504 0.7232 0.6712 0.7256 0.7772 0.0000 0.0000 0.0000}
	\noindent\resizebox{\linewidth}{!}{0.6648 0.5928 0.6204 0.6936 0.6848 0.7172 0.6624 0.6900 0.7308 0.6484 0.7168 0.6424 0.7336 0.7672 0.6684 0.0000 0.0000}
	\noindent\resizebox{\linewidth}{!}{0.6704 0.5920 0.6152 0.6984 0.6988 0.7260 0.6732 0.6764 0.7260 0.6464 0.7064 0.6292 0.7216 0.7520 0.6780 0.7612 0.0000}
	\noindent\resizebox{\linewidth}{!}{0.6736 0.6140 0.6168 0.6928 0.6996 0.7228 0.6792 0.6908 0.7284 0.6736 0.7136 0.6452 0.7208 0.7496 0.6492 0.7416 0.7920}

\end{spacing}

\paragraph{MEGA:}

~

\begin{spacing}{.5}
	
	\noindent\resizebox{\linewidth}{!}{0.6472 0.0000 0.0000 0.0000 0.0000 0.0000 0.0000 0.0000 0.0000 0.0000 0.0000 0.0000 0.0000 0.0000 0.0000 0.0000 0.0000}
	
	\noindent\resizebox{\linewidth}{!}{0.6260 0.5824 0.0000 0.0000 0.0000 0.0000 0.0000 0.0000 0.0000 0.0000 0.0000 0.0000 0.0000 0.0000 0.0000 0.0000 0.0000}
	
	\noindent\resizebox{\linewidth}{!}{0.6324 0.5700 0.6300 0.0000 0.0000 0.0000 0.0000 0.0000 0.0000 0.0000 0.0000 0.0000 0.0000 0.0000 0.0000 0.0000 0.0000}
	
	\noindent\resizebox{\linewidth}{!}{0.6236 0.5496 0.5624 0.6452 0.0000 0.0000 0.0000 0.0000 0.0000 0.0000 0.0000 0.0000 0.0000 0.0000 0.0000 0.0000 0.0000}
	
	\noindent\resizebox{\linewidth}{!}{0.6132 0.5612 0.6048 0.6736 0.6960 0.0000 0.0000 0.0000 0.0000 0.0000 0.0000 0.0000 0.0000 0.0000 0.0000 0.0000 0.0000}
	
	\noindent\resizebox{\linewidth}{!}{0.6140 0.5628 0.5692 0.6632 0.6792 0.7688 0.0000 0.0000 0.0000 0.0000 0.0000 0.0000 0.0000 0.0000 0.0000 0.0000 0.0000}
	
	\noindent\resizebox{\linewidth}{!}{0.5780 0.5500 0.5864 0.6364 0.6792 0.7420 0.6868 0.0000 0.0000 0.0000 0.0000 0.0000 0.0000 0.0000 0.0000 0.0000 0.0000}
	
	\noindent\resizebox{\linewidth}{!}{0.5756 0.5308 0.5764 0.6292 0.6500 0.6820 0.6580 0.6580 0.0000 0.0000 0.0000 0.0000 0.0000 0.0000 0.0000 0.0000 0.0000}
	
	\noindent\resizebox{\linewidth}{!}{0.6212 0.5704 0.5876 0.6376 0.6600 0.7056 0.6636 0.6916 0.7376 0.0000 0.0000 0.0000 0.0000 0.0000 0.0000 0.0000 0.0000}
	
	\noindent\resizebox{\linewidth}{!}{0.5992 0.5580 0.5828 0.6212 0.6528 0.6512 0.6496 0.6700 0.7276 0.6732 0.0000 0.0000 0.0000 0.0000 0.0000 0.0000 0.0000}
	
	\noindent\resizebox{\linewidth}{!}{0.6104 0.5552 0.5804 0.6396 0.6700 0.6960 0.6568 0.6752 0.7412 0.6752 0.7432 0.0000 0.0000 0.0000 0.0000 0.0000 0.0000}
	
	\noindent\resizebox{\linewidth}{!}{0.5880 0.5516 0.5816 0.6248 0.6552 0.6568 0.6412 0.6284 0.7044 0.6304 0.7196 0.6660 0.0000 0.0000 0.0000 0.0000 0.0000}
	
	\noindent\resizebox{\linewidth}{!}{0.6020 0.5628 0.5792 0.6164 0.6444 0.6636 0.6356 0.6536 0.7008 0.6132 0.7000 0.6336 0.7108 0.0000 0.0000 0.0000 0.0000}
	
	\noindent\resizebox{\linewidth}{!}{0.6124 0.5692 0.5924 0.6420 0.6516 0.6912 0.6352 0.6492 0.6848 0.6400 0.6872 0.6312 0.7280 0.7596 0.0000 0.0000 0.0000}
	
	\noindent\resizebox{\linewidth}{!}{0.6012 0.5468 0.5908 0.6128 0.6552 0.6852 0.6288 0.6428 0.6704 0.6176 0.6988 0.6268 0.7088 0.7348 0.6324 0.0000 0.0000}
	
	\noindent\resizebox{\linewidth}{!}{0.6244 0.5588 0.5960 0.6432 0.6448 0.6868 0.6388 0.6540 0.6896 0.6056 0.6920 0.6192 0.7124 0.7344 0.6560 0.7604 0.0000}
	
	\noindent\resizebox{\linewidth}{!}{0.6088 0.5896 0.5840 0.6552 0.6716 0.6904 0.6584 0.6372 0.7032 0.6300 0.6900 0.5864 0.6832 0.7140 0.6348 0.7264 0.7780}
	
\end{spacing}

\paragraph{A-GEM:}

~

\begin{spacing}{.5}
	
	\noindent\resizebox{\linewidth}{!}{0.6772 0.0000 0.0000 0.0000 0.0000 0.0000 0.0000 0.0000 0.0000 0.0000 0.0000 0.0000 0.0000 0.0000 0.0000 0.0000 0.0000}
	
	\noindent\resizebox{\linewidth}{!}{0.5948 0.5764 0.0000 0.0000 0.0000 0.0000 0.0000 0.0000 0.0000 0.0000 0.0000 0.0000 0.0000 0.0000 0.0000 0.0000 0.0000}
	\noindent\resizebox{\linewidth}{!}{0.6324 0.5828 0.6432 0.0000 0.0000 0.0000 0.0000 0.0000 0.0000 0.0000 0.0000 0.0000 0.0000 0.0000 0.0000 0.0000 0.0000}
	\noindent\resizebox{\linewidth}{!}{0.5980 0.5384 0.5396 0.6456 0.0000 0.0000 0.0000 0.0000 0.0000 0.0000 0.0000 0.0000 0.0000 0.0000 0.0000 0.0000 0.0000}
	\noindent\resizebox{\linewidth}{!}{0.5864 0.5404 0.5576 0.6436 0.7004 0.0000 0.0000 0.0000 0.0000 0.0000 0.0000 0.0000 0.0000 0.0000 0.0000 0.0000 0.0000}
	\noindent\resizebox{\linewidth}{!}{0.5728 0.5392 0.5068 0.5940 0.6344 0.7180 0.0000 0.0000 0.0000 0.0000 0.0000 0.0000 0.0000 0.0000 0.0000 0.0000 0.0000}
	\noindent\resizebox{\linewidth}{!}{0.5572 0.5404 0.5308 0.6224 0.6116 0.6520 0.6688 0.0000 0.0000 0.0000 0.0000 0.0000 0.0000 0.0000 0.0000 0.0000 0.0000}
	\noindent\resizebox{\linewidth}{!}{0.6064 0.5356 0.5492 0.5872 0.6164 0.6532 0.6296 0.6724 0.0000 0.0000 0.0000 0.0000 0.0000 0.0000 0.0000 0.0000 0.0000}
	\noindent\resizebox{\linewidth}{!}{0.6060 0.5472 0.5528 0.6236 0.5920 0.6348 0.6076 0.6348 0.6972 0.0000 0.0000 0.0000 0.0000 0.0000 0.0000 0.0000 0.0000}
	\noindent\resizebox{\linewidth}{!}{0.6004 0.5080 0.4960 0.6128 0.5656 0.6356 0.5752 0.6140 0.6580 0.6792 0.0000 0.0000 0.0000 0.0000 0.0000 0.0000 0.0000}
	\noindent\resizebox{\linewidth}{!}{0.5992 0.5408 0.5332 0.5964 0.5928 0.6520 0.5928 0.6304 0.6764 0.5916 0.7364 0.0000 0.0000 0.0000 0.0000 0.0000 0.0000}
	\noindent\resizebox{\linewidth}{!}{0.5748 0.5020 0.5104 0.5936 0.6016 0.6184 0.5772 0.6164 0.6408 0.5688 0.6776 0.6436 0.0000 0.0000 0.0000 0.0000 0.0000}
	\noindent\resizebox{\linewidth}{!}{0.6056 0.5100 0.5200 0.5916 0.6012 0.6056 0.5816 0.6060 0.6236 0.5808 0.6288 0.5768 0.7332 0.0000 0.0000 0.0000 0.0000}
	\noindent\resizebox{\linewidth}{!}{0.6184 0.5344 0.5308 0.5888 0.6116 0.6188 0.6012 0.6248 0.6136 0.5836 0.6428 0.5688 0.6524 0.7392 0.0000 0.0000 0.0000}
	\noindent\resizebox{\linewidth}{!}{0.6012 0.5220 0.5488 0.6008 0.5828 0.6048 0.5728 0.5884 0.6356 0.5740 0.6476 0.5540 0.6520 0.6804 0.6460 0.0000 0.0000}
	\noindent\resizebox{\linewidth}{!}{0.5984 0.5360 0.5520 0.5808 0.5704 0.6184 0.6068 0.6108 0.6452 0.5404 0.6520 0.5256 0.6624 0.6512 0.5864 0.7388 0.0000}
	\noindent\resizebox{\linewidth}{!}{0.6232 0.5356 0.5412 0.6104 0.6080 0.6248 0.5944 0.5900 0.6492 0.5872 0.6468 0.5352 0.6336 0.6520 0.5908 0.6444 0.7508}
	
\end{spacing}

\subsubsection{SPLIT CUB}

\paragraph{MDMT-R:}

~

\begin{spacing}{.5}
	
	\noindent\resizebox{\linewidth}{!}{0.4895 0.0000 0.0000 0.0000 0.0000 0.0000 0.0000 0.0000 0.0000 0.0000 0.0000 0.0000 0.0000 0.0000 0.0000 0.0000 0.0000}
	\noindent\resizebox{\linewidth}{!}{0.6745 0.5760 0.0000 0.0000 0.0000 0.0000 0.0000 0.0000 0.0000 0.0000 0.0000 0.0000 0.0000 0.0000 0.0000 0.0000 0.0000}
	\noindent\resizebox{\linewidth}{!}{0.7401 0.7400 0.6662 0.0000 0.0000 0.0000 0.0000 0.0000 0.0000 0.0000 0.0000 0.0000 0.0000 0.0000 0.0000 0.0000 0.0000}
	\noindent\resizebox{\linewidth}{!}{0.7442 0.7663 0.7934 0.7369 0.0000 0.0000 0.0000 0.0000 0.0000 0.0000 0.0000 0.0000 0.0000 0.0000 0.0000 0.0000 0.0000}
	\noindent\resizebox{\linewidth}{!}{0.7583 0.7702 0.7941 0.8119 0.6626 0.0000 0.0000 0.0000 0.0000 0.0000 0.0000 0.0000 0.0000 0.0000 0.0000 0.0000 0.0000}
	\noindent\resizebox{\linewidth}{!}{0.7710 0.7837 0.7941 0.7424 0.7669 0.7327 0.0000 0.0000 0.0000 0.0000 0.0000 0.0000 0.0000 0.0000 0.0000 0.0000 0.0000}
	\noindent\resizebox{\linewidth}{!}{0.7811 0.7697 0.8204 0.8044 0.7630 0.8067 0.7581 0.0000 0.0000 0.0000 0.0000 0.0000 0.0000 0.0000 0.0000 0.0000 0.0000}
	\noindent\resizebox{\linewidth}{!}{0.7971 0.7810 0.8303 0.8311 0.8019 0.8095 0.8337 0.7485 0.0000 0.0000 0.0000 0.0000 0.0000 0.0000 0.0000 0.0000 0.0000}
	\noindent\resizebox{\linewidth}{!}{0.7703 0.8008 0.8249 0.8081 0.7980 0.8369 0.8662 0.7686 0.7511 0.0000 0.0000 0.0000 0.0000 0.0000 0.0000 0.0000 0.0000}
	\noindent\resizebox{\linewidth}{!}{0.8047 0.8215 0.8316 0.8165 0.8089 0.8049 0.8463 0.7734 0.7968 0.7640 0.0000 0.0000 0.0000 0.0000 0.0000 0.0000 0.0000}
	\noindent\resizebox{\linewidth}{!}{0.7924 0.8011 0.8303 0.8435 0.8126 0.8358 0.8531 0.7990 0.7823 0.8107 0.8219 0.0000 0.0000 0.0000 0.0000 0.0000 0.0000}
	\noindent\resizebox{\linewidth}{!}{0.7886 0.7963 0.8114 0.8250 0.7875 0.8222 0.8249 0.8164 0.7612 0.8091 0.8305 0.7917 0.0000 0.0000 0.0000 0.0000 0.0000}
	\noindent\resizebox{\linewidth}{!}{0.8000 0.8034 0.8516 0.8224 0.8052 0.8355 0.8185 0.7993 0.7778 0.8177 0.8461 0.8107 0.8405 0.0000 0.0000 0.0000 0.0000}
	\noindent\resizebox{\linewidth}{!}{0.8276 0.8085 0.8549 0.8120 0.8289 0.8487 0.8130 0.8041 0.7675 0.8183 0.8329 0.8060 0.8332 0.7764 0.0000 0.0000 0.0000}
	\noindent\resizebox{\linewidth}{!}{0.7938 0.8199 0.8154 0.8069 0.8141 0.8358 0.8395 0.7959 0.8041 0.8286 0.8365 0.8043 0.8527 0.7462 0.8142 0.0000 0.0000}
	\noindent\resizebox{\linewidth}{!}{0.8215 0.7984 0.8517 0.8471 0.8237 0.8352 0.8538 0.8115 0.7959 0.8302 0.8366 0.8304 0.8494 0.8036 0.8071 0.8064 0.0000}
	\noindent\resizebox{\linewidth}{!}{0.8342 0.8331 0.8544 0.8695 0.8449 0.8628 0.8789 0.8328 0.8235 0.8601 0.8938 0.8434 0.8538 0.8096 0.8195 0.8166 0.7844}

\end{spacing}

\paragraph{MEGA:}

~

\begin{spacing}{.5}
	
	\noindent\resizebox{\linewidth}{!}{0.4050 0.0000 0.0000 0.0000 0.0000 0.0000 0.0000 0.0000 0.0000 0.0000 0.0000 0.0000 0.0000 0.0000 0.0000 0.0000 0.0000}
	\noindent\resizebox{\linewidth}{!}{0.6876 0.5088 0.0000 0.0000 0.0000 0.0000 0.0000 0.0000 0.0000 0.0000 0.0000 0.0000 0.0000 0.0000 0.0000 0.0000 0.0000}
	\noindent\resizebox{\linewidth}{!}{0.6844 0.7244 0.6263 0.0000 0.0000 0.0000 0.0000 0.0000 0.0000 0.0000 0.0000 0.0000 0.0000 0.0000 0.0000 0.0000 0.0000}
	\noindent\resizebox{\linewidth}{!}{0.7071 0.7445 0.7753 0.6553 0.0000 0.0000 0.0000 0.0000 0.0000 0.0000 0.0000 0.0000 0.0000 0.0000 0.0000 0.0000 0.0000}
	\noindent\resizebox{\linewidth}{!}{0.7294 0.7483 0.7941 0.7787 0.6857 0.0000 0.0000 0.0000 0.0000 0.0000 0.0000 0.0000 0.0000 0.0000 0.0000 0.0000 0.0000}
	\noindent\resizebox{\linewidth}{!}{0.7049 0.7266 0.7712 0.7538 0.7528 0.6366 0.0000 0.0000 0.0000 0.0000 0.0000 0.0000 0.0000 0.0000 0.0000 0.0000 0.0000}
	\noindent\resizebox{\linewidth}{!}{0.7271 0.7488 0.7877 0.7893 0.7861 0.7905 0.6830 0.0000 0.0000 0.0000 0.0000 0.0000 0.0000 0.0000 0.0000 0.0000 0.0000}
	\noindent\resizebox{\linewidth}{!}{0.7316 0.7737 0.7920 0.7867 0.7851 0.7956 0.8179 0.6816 0.0000 0.0000 0.0000 0.0000 0.0000 0.0000 0.0000 0.0000 0.0000}
	\noindent\resizebox{\linewidth}{!}{0.7503 0.7525 0.7860 0.7947 0.7838 0.7818 0.8148 0.7912 0.6934 0.0000 0.0000 0.0000 0.0000 0.0000 0.0000 0.0000 0.0000}
	\noindent\resizebox{\linewidth}{!}{0.7445 0.7475 0.7819 0.7599 0.7680 0.7834 0.7901 0.7858 0.7883 0.7085 0.0000 0.0000 0.0000 0.0000 0.0000 0.0000 0.0000}
	\noindent\resizebox{\linewidth}{!}{0.7506 0.7732 0.8053 0.8129 0.8049 0.8099 0.8275 0.7938 0.8041 0.8146 0.7211 0.0000 0.0000 0.0000 0.0000 0.0000 0.0000}
	\noindent\resizebox{\linewidth}{!}{0.7475 0.7670 0.7860 0.7982 0.7818 0.7892 0.8133 0.7881 0.8014 0.8145 0.7958 0.7144 0.0000 0.0000 0.0000 0.0000 0.0000}
	\noindent\resizebox{\linewidth}{!}{0.7393 0.7572 0.7907 0.8006 0.7821 0.8043 0.8123 0.7629 0.7905 0.7902 0.8161 0.7622 0.7262 0.0000 0.0000 0.0000 0.0000}
	\noindent\resizebox{\linewidth}{!}{0.7492 0.7501 0.8031 0.8098 0.7956 0.8020 0.8053 0.7811 0.7873 0.8081 0.8159 0.7672 0.7981 0.6794 0.0000 0.0000 0.0000}
	\noindent\resizebox{\linewidth}{!}{0.7488 0.7661 0.7958 0.7830 0.7845 0.7873 0.8007 0.7759 0.7902 0.8139 0.8142 0.7746 0.7984 0.7651 0.6831 0.0000 0.0000}
	\noindent\resizebox{\linewidth}{!}{0.7538 0.7857 0.8116 0.8170 0.8050 0.8052 0.8277 0.8024 0.8070 0.8220 0.8301 0.7767 0.8210 0.7839 0.7777 0.7505 0.0000}
	\noindent\resizebox{\linewidth}{!}{0.7728 0.7846 0.8203 0.7995 0.8141 0.8219 0.8201 0.7973 0.8109 0.8366 0.8390 0.7880 0.8300 0.7948 0.8078 0.8112 0.7418}
	
\end{spacing}

\paragraph{A-GEM:}

~

\begin{spacing}{.5}
	
	\noindent\resizebox{\linewidth}{!}{0.4263 0.0000 0.0000 0.0000 0.0000 0.0000 0.0000 0.0000 0.0000 0.0000 0.0000 0.0000 0.0000 0.0000 0.0000 0.0000 0.0000}
	\noindent\resizebox{\linewidth}{!}{0.4383 0.5243 0.0000 0.0000 0.0000 0.0000 0.0000 0.0000 0.0000 0.0000 0.0000 0.0000 0.0000 0.0000 0.0000 0.0000 0.0000}
	\noindent\resizebox{\linewidth}{!}{0.4642 0.5220 0.6064 0.0000 0.0000 0.0000 0.0000 0.0000 0.0000 0.0000 0.0000 0.0000 0.0000 0.0000 0.0000 0.0000 0.0000}
	\noindent\resizebox{\linewidth}{!}{0.4850 0.5420 0.6057 0.6765 0.0000 0.0000 0.0000 0.0000 0.0000 0.0000 0.0000 0.0000 0.0000 0.0000 0.0000 0.0000 0.0000}
	\noindent\resizebox{\linewidth}{!}{0.4935 0.5378 0.6328 0.6042 0.6621 0.0000 0.0000 0.0000 0.0000 0.0000 0.0000 0.0000 0.0000 0.0000 0.0000 0.0000 0.0000}
	\noindent\resizebox{\linewidth}{!}{0.4678 0.5071 0.6369 0.5585 0.5966 0.6137 0.0000 0.0000 0.0000 0.0000 0.0000 0.0000 0.0000 0.0000 0.0000 0.0000 0.0000}
	\noindent\resizebox{\linewidth}{!}{0.4909 0.5630 0.6435 0.6369 0.6140 0.6278 0.6825 0.0000 0.0000 0.0000 0.0000 0.0000 0.0000 0.0000 0.0000 0.0000 0.0000}
	\noindent\resizebox{\linewidth}{!}{0.4928 0.5607 0.6157 0.5992 0.6025 0.6037 0.6572 0.6617 0.0000 0.0000 0.0000 0.0000 0.0000 0.0000 0.0000 0.0000 0.0000}
	\noindent\resizebox{\linewidth}{!}{0.4850 0.5401 0.6090 0.5830 0.6004 0.5980 0.6384 0.6520 0.6850 0.0000 0.0000 0.0000 0.0000 0.0000 0.0000 0.0000 0.0000}
	\noindent\resizebox{\linewidth}{!}{0.4901 0.5542 0.6086 0.5858 0.6080 0.5985 0.6206 0.6533 0.6607 0.6964 0.0000 0.0000 0.0000 0.0000 0.0000 0.0000 0.0000}
	\noindent\resizebox{\linewidth}{!}{0.4909 0.5455 0.6465 0.6220 0.6241 0.6186 0.6094 0.6197 0.6181 0.6569 0.7128 0.0000 0.0000 0.0000 0.0000 0.0000 0.0000}
	\noindent\resizebox{\linewidth}{!}{0.5075 0.5675 0.6266 0.6126 0.5999 0.6037 0.6073 0.6214 0.6004 0.6424 0.6547 0.6673 0.0000 0.0000 0.0000 0.0000 0.0000}
	\noindent\resizebox{\linewidth}{!}{0.4909 0.5588 0.5943 0.5852 0.5898 0.6018 0.6188 0.5974 0.6266 0.6228 0.6792 0.6281 0.7035 0.0000 0.0000 0.0000 0.0000}
	\noindent\resizebox{\linewidth}{!}{0.5297 0.5535 0.6220 0.6281 0.6383 0.6037 0.6234 0.5967 0.6277 0.6051 0.6393 0.5950 0.6965 0.6630 0.0000 0.0000 0.0000}
	\noindent\resizebox{\linewidth}{!}{0.5104 0.5677 0.6446 0.6198 0.6135 0.6093 0.6100 0.5710 0.5912 0.6314 0.6312 0.5951 0.6924 0.6422 0.6584 0.0000 0.0000}
	\noindent\resizebox{\linewidth}{!}{0.5196 0.5527 0.6350 0.6043 0.6384 0.6009 0.6226 0.5971 0.6028 0.6210 0.6507 0.6318 0.6757 0.6003 0.5899 0.7272 0.0000}
	\noindent\resizebox{\linewidth}{!}{0.5345 0.5519 0.6395 0.6064 0.6335 0.6000 0.5879 0.5899 0.5869 0.6509 0.6486 0.6332 0.6914 0.6130 0.5983 0.6806 0.7216}
	
\end{spacing}

\subsubsection{SPLIT AWA}

\paragraph{MDMT-R:}

~

\begin{spacing}{.5}
	
	\noindent\resizebox{\linewidth}{!}{0.3792 0.0000 0.0000 0.0000 0.0000 0.0000 0.0000 0.0000 0.0000 0.0000 0.0000 0.0000 0.0000 0.0000 0.0000 0.0000 0.0000}
	\noindent\resizebox{\linewidth}{!}{0.5119 0.4335 0.0000 0.0000 0.0000 0.0000 0.0000 0.0000 0.0000 0.0000 0.0000 0.0000 0.0000 0.0000 0.0000 0.0000 0.0000}
	\noindent\resizebox{\linewidth}{!}{0.5147 0.4985 0.4556 0.0000 0.0000 0.0000 0.0000 0.0000 0.0000 0.0000 0.0000 0.0000 0.0000 0.0000 0.0000 0.0000 0.0000}
	\noindent\resizebox{\linewidth}{!}{0.5382 0.5230 0.5662 0.3475 0.0000 0.0000 0.0000 0.0000 0.0000 0.0000 0.0000 0.0000 0.0000 0.0000 0.0000 0.0000 0.0000}
	\noindent\resizebox{\linewidth}{!}{0.5292 0.5307 0.5874 0.4767 0.4884 0.0000 0.0000 0.0000 0.0000 0.0000 0.0000 0.0000 0.0000 0.0000 0.0000 0.0000 0.0000}
	\noindent\resizebox{\linewidth}{!}{0.5725 0.5134 0.5924 0.5043 0.5974 0.4749 0.0000 0.0000 0.0000 0.0000 0.0000 0.0000 0.0000 0.0000 0.0000 0.0000 0.0000}
	\noindent\resizebox{\linewidth}{!}{0.5690 0.5398 0.6020 0.4712 0.5798 0.6075 0.4768 0.0000 0.0000 0.0000 0.0000 0.0000 0.0000 0.0000 0.0000 0.0000 0.0000}
	\noindent\resizebox{\linewidth}{!}{0.5645 0.5055 0.6122 0.4804 0.5792 0.5996 0.5433 0.5065 0.0000 0.0000 0.0000 0.0000 0.0000 0.0000 0.0000 0.0000 0.0000}
	\noindent\resizebox{\linewidth}{!}{0.5561 0.5574 0.6216 0.5301 0.6216 0.6023 0.5638 0.5725 0.4626 0.0000 0.0000 0.0000 0.0000 0.0000 0.0000 0.0000 0.0000}
	\noindent\resizebox{\linewidth}{!}{0.5941 0.5437 0.6306 0.5239 0.6218 0.6206 0.5343 0.5575 0.5019 0.5462 0.0000 0.0000 0.0000 0.0000 0.0000 0.0000 0.0000}
	\noindent\resizebox{\linewidth}{!}{0.6117 0.5522 0.6058 0.5104 0.6036 0.6073 0.5348 0.5694 0.5112 0.6453 0.5159 0.0000 0.0000 0.0000 0.0000 0.0000 0.0000}
	\noindent\resizebox{\linewidth}{!}{0.5784 0.5168 0.5892 0.5313 0.6071 0.5785 0.5446 0.5509 0.5224 0.6316 0.5863 0.5257 0.0000 0.0000 0.0000 0.0000 0.0000}
	\noindent\resizebox{\linewidth}{!}{0.5837 0.5408 0.6202 0.5282 0.6289 0.6234 0.5625 0.5494 0.5276 0.6285 0.6178 0.6038 0.4910 0.0000 0.0000 0.0000 0.0000}
	\noindent\resizebox{\linewidth}{!}{0.5962 0.5716 0.6666 0.5407 0.6461 0.6104 0.5786 0.5918 0.5276 0.6605 0.6201 0.6079 0.5582 0.5378 0.0000 0.0000 0.0000}
	\noindent\resizebox{\linewidth}{!}{0.6177 0.5444 0.6474 0.5467 0.6466 0.6326 0.5556 0.6075 0.5505 0.6552 0.6383 0.6123 0.6076 0.6551 0.5770 0.0000 0.0000}
	\noindent\resizebox{\linewidth}{!}{0.5900 0.5438 0.6428 0.5373 0.6405 0.5971 0.5792 0.5991 0.5428 0.6470 0.6332 0.5935 0.5679 0.6485 0.6678 0.5730 0.0000}
	\noindent\resizebox{\linewidth}{!}{0.6184 0.5875 0.6696 0.5620 0.6434 0.6194 0.5766 0.6226 0.5566 0.6739 0.6475 0.6349 0.5714 0.6531 0.6640 0.6051 0.5604}
	
\end{spacing}

\paragraph{MEGA:}

~

\begin{spacing}{.5}
	
	\noindent\resizebox{\linewidth}{!}{0.4101 0.0000 0.0000 0.0000 0.0000 0.0000 0.0000 0.0000 0.0000 0.0000 0.0000 0.0000 0.0000 0.0000 0.0000 0.0000 0.0000}
	\noindent\resizebox{\linewidth}{!}{0.4791 0.4377 0.0000 0.0000 0.0000 0.0000 0.0000 0.0000 0.0000 0.0000 0.0000 0.0000 0.0000 0.0000 0.0000 0.0000 0.0000}
	\noindent\resizebox{\linewidth}{!}{0.5061 0.5174 0.4116 0.0000 0.0000 0.0000 0.0000 0.0000 0.0000 0.0000 0.0000 0.0000 0.0000 0.0000 0.0000 0.0000 0.0000}
	\noindent\resizebox{\linewidth}{!}{0.5184 0.5219 0.4962 0.4741 0.0000 0.0000 0.0000 0.0000 0.0000 0.0000 0.0000 0.0000 0.0000 0.0000 0.0000 0.0000 0.0000}
	\noindent\resizebox{\linewidth}{!}{0.4931 0.5219 0.4828 0.5142 0.4026 0.0000 0.0000 0.0000 0.0000 0.0000 0.0000 0.0000 0.0000 0.0000 0.0000 0.0000 0.0000}
	\noindent\resizebox{\linewidth}{!}{0.5122 0.5357 0.4894 0.5293 0.5053 0.4680 0.0000 0.0000 0.0000 0.0000 0.0000 0.0000 0.0000 0.0000 0.0000 0.0000 0.0000}
	\noindent\resizebox{\linewidth}{!}{0.5242 0.5226 0.5038 0.5219 0.5215 0.5605 0.4820 0.0000 0.0000 0.0000 0.0000 0.0000 0.0000 0.0000 0.0000 0.0000 0.0000}
	\noindent\resizebox{\linewidth}{!}{0.5383 0.5359 0.5216 0.5323 0.5204 0.5551 0.5774 0.4682 0.0000 0.0000 0.0000 0.0000 0.0000 0.0000 0.0000 0.0000 0.0000}
	\noindent\resizebox{\linewidth}{!}{0.5291 0.5179 0.4850 0.5367 0.5043 0.5380 0.5565 0.5111 0.4698 0.0000 0.0000 0.0000 0.0000 0.0000 0.0000 0.0000 0.0000}
	\noindent\resizebox{\linewidth}{!}{0.5386 0.5106 0.4995 0.5245 0.5150 0.5470 0.5457 0.4862 0.5380 0.4619 0.0000 0.0000 0.0000 0.0000 0.0000 0.0000 0.0000}
	\noindent\resizebox{\linewidth}{!}{0.5612 0.5569 0.5260 0.5566 0.5387 0.5765 0.5847 0.5483 0.5444 0.5622 0.5542 0.0000 0.0000 0.0000 0.0000 0.0000 0.0000}
	\noindent\resizebox{\linewidth}{!}{0.5593 0.5593 0.5416 0.5611 0.5345 0.5348 0.5539 0.5358 0.5418 0.5363 0.5720 0.4725 0.0000 0.0000 0.0000 0.0000 0.0000}
	\noindent\resizebox{\linewidth}{!}{0.5471 0.5522 0.5333 0.5383 0.5159 0.5410 0.5638 0.5234 0.5234 0.5342 0.5614 0.5611 0.5026 0.0000 0.0000 0.0000 0.0000}
	\noindent\resizebox{\linewidth}{!}{0.5557 0.5561 0.5391 0.5368 0.5093 0.5442 0.5743 0.5026 0.5484 0.5126 0.5788 0.5548 0.5641 0.5386 0.0000 0.0000 0.0000}
	\noindent\resizebox{\linewidth}{!}{0.5240 0.5563 0.5366 0.5465 0.5102 0.5659 0.5688 0.5071 0.5320 0.5183 0.5865 0.5705 0.5295 0.6206 0.5483 0.0000 0.0000}
	\noindent\resizebox{\linewidth}{!}{0.5541 0.5575 0.5219 0.5559 0.5310 0.5490 0.5983 0.5121 0.5466 0.5204 0.5929 0.5626 0.5466 0.6064 0.6291 0.4661 0.0000}
	\noindent\resizebox{\linewidth}{!}{0.5542 0.5469 0.5093 0.5677 0.5136 0.5471 0.5581 0.4954 0.5326 0.5169 0.5857 0.5771 0.5345 0.5964 0.6081 0.5187 0.4655}
	
\end{spacing}

\paragraph{A-GEM:}

~

\begin{spacing}{.5}
	
	\noindent\resizebox{\linewidth}{!}{0.4127 0.0000 0.0000 0.0000 0.0000 0.0000 0.0000 0.0000 0.0000 0.0000 0.0000 0.0000 0.0000 0.0000 0.0000 0.0000 0.0000}
	\noindent\resizebox{\linewidth}{!}{0.4256 0.4422 0.0000 0.0000 0.0000 0.0000 0.0000 0.0000 0.0000 0.0000 0.0000 0.0000 0.0000 0.0000 0.0000 0.0000 0.0000}
	\noindent\resizebox{\linewidth}{!}{0.4436 0.4445 0.4058 0.0000 0.0000 0.0000 0.0000 0.0000 0.0000 0.0000 0.0000 0.0000 0.0000 0.0000 0.0000 0.0000 0.0000}
	\noindent\resizebox{\linewidth}{!}{0.4371 0.4784 0.4334 0.4463 0.0000 0.0000 0.0000 0.0000 0.0000 0.0000 0.0000 0.0000 0.0000 0.0000 0.0000 0.0000 0.0000}
	\noindent\resizebox{\linewidth}{!}{0.4339 0.4795 0.4236 0.4258 0.3963 0.0000 0.0000 0.0000 0.0000 0.0000 0.0000 0.0000 0.0000 0.0000 0.0000 0.0000 0.0000}
	\noindent\resizebox{\linewidth}{!}{0.4226 0.4674 0.4311 0.4505 0.3864 0.4495 0.0000 0.0000 0.0000 0.0000 0.0000 0.0000 0.0000 0.0000 0.0000 0.0000 0.0000}
	\noindent\resizebox{\linewidth}{!}{0.4279 0.4462 0.4268 0.4217 0.3854 0.4254 0.4239 0.0000 0.0000 0.0000 0.0000 0.0000 0.0000 0.0000 0.0000 0.0000 0.0000}
	\noindent\resizebox{\linewidth}{!}{0.4621 0.4733 0.4421 0.4563 0.4239 0.4356 0.4489 0.4299 0.0000 0.0000 0.0000 0.0000 0.0000 0.0000 0.0000 0.0000 0.0000}
	\noindent\resizebox{\linewidth}{!}{0.4672 0.4774 0.4363 0.4402 0.4265 0.4387 0.4503 0.4129 0.4431 0.0000 0.0000 0.0000 0.0000 0.0000 0.0000 0.0000 0.0000}
	\noindent\resizebox{\linewidth}{!}{0.4659 0.4417 0.4385 0.4386 0.4188 0.4419 0.4655 0.4075 0.3971 0.4286 0.0000 0.0000 0.0000 0.0000 0.0000 0.0000 0.0000}
	\noindent\resizebox{\linewidth}{!}{0.4555 0.4657 0.4495 0.4645 0.4133 0.4374 0.4717 0.4083 0.4289 0.4144 0.5037 0.0000 0.0000 0.0000 0.0000 0.0000 0.0000}
	\noindent\resizebox{\linewidth}{!}{0.4425 0.4463 0.4277 0.4532 0.4301 0.4319 0.4865 0.4288 0.4043 0.4021 0.4172 0.4478 0.0000 0.0000 0.0000 0.0000 0.0000}
	\noindent\resizebox{\linewidth}{!}{0.4427 0.4582 0.4438 0.4527 0.4345 0.4638 0.4895 0.4327 0.4181 0.4295 0.4745 0.4303 0.4889 0.0000 0.0000 0.0000 0.0000}
	\noindent\resizebox{\linewidth}{!}{0.4538 0.4519 0.4066 0.4696 0.4099 0.4459 0.4859 0.4080 0.3931 0.3827 0.4382 0.3854 0.4055 0.4736 0.0000 0.0000 0.0000}
	\noindent\resizebox{\linewidth}{!}{0.4476 0.4786 0.4148 0.4879 0.4269 0.4576 0.5014 0.4532 0.4264 0.4137 0.4535 0.4178 0.3955 0.4809 0.4941 0.0000 0.0000}
	\noindent\resizebox{\linewidth}{!}{0.4482 0.4635 0.4114 0.4720 0.4231 0.4527 0.5082 0.4076 0.4231 0.4262 0.4436 0.4122 0.3915 0.4809 0.4762 0.4286 0.0000}
	\noindent\resizebox{\linewidth}{!}{0.4530 0.4605 0.4294 0.4519 0.4388 0.4649 0.4885 0.4514 0.4395 0.4236 0.4760 0.4495 0.4105 0.4621 0.4730 0.4039 0.4646}
	
\end{spacing}

\end{document}